\documentclass[letterpaper, 10 pt, journal, twoside]{ieeetran}  

\usepackage{amsmath}

\usepackage{amsthm}
\usepackage[font=footnotesize,labelfont=normalfont]{caption}
\usepackage{subcaption}
\usepackage{xcolor}
\usepackage{hyperref}
\hypersetup{
    colorlinks=true,
    linkcolor=blue,
    filecolor=magenta,      
    urlcolor=cyan,
    pdftitle={Overleaf Example},
    pdfpagemode=FullScreen,
    }
\usepackage{algpseudocode}
\usepackage[linesnumbered,ruled,vlined]{algorithm2e}
\usepackage{amsfonts}
\usepackage{svg}
\usepackage{cite}
\usepackage{siunitx}
\usepackage{upgreek}

\DeclareMathOperator*{\argmin}{arg\,min}
\newcommand{\norm}[1]{\left\lVert#1\right\rVert}
\theoremstyle{definition}
\newtheorem{definition}{Definition}[]
\newtheorem{theorem}{Theorem}

\usepackage{bbm}
\usepackage[most]{tcolorbox}
\newcommand{\newtext}[1]{{\color{black}{{#1}}}}
\tcbset{
  highlightbox/.style={
    colback=blue!5!white,    
    colframe=blue!10!white,  
    boxrule=0pt,             
    arc=2mm,                 
    left=2mm, right=2mm, top=1mm, bottom=1mm,
    enhanced
  }
}
\def\IEEEtitletopspace{3pt}
\usepackage[top=57pt,bottom=43pt,left=48pt,right=48pt]{geometry}

\title{\LARGE \bf
Online Learning-Enhanced High Order Adaptive Safety Control
}

\author{Lishuo Pan$^{1}$, Mattia Catellani$^{2}$, Thales C. Silva$^{1}$, Lorenzo Sabattini$^{2}$, Nora Ayanian$^{1}$%
\thanks{Manuscript received November 20, 2025; Revised February 26, 2026; Accepted March 30, 2026. 
This paper was recommended for publication by Editor Tamim Asfour and Editor Cl\'ement Gosselin upon evaluation of the Associate Editor and Reviewers' comments.
This work was supported by NSF grants 2317145, 2311967, 2330942.} 
\thanks{$^{1}$Lishuo Pan, Thales C. Silva and Nora Ayanian are with the Department of Computer Science, Brown University, Providence, RI 02912 USA. 
        Email: {\tt\footnotesize \{lishuo\_pan, thales\_silva, nora\_ayanian\}@brown.edu}.}\\ %
\thanks{$^{2}$Mattia Catellani and Lorenzo Sabattini are with the Department of Sciences and Methods for Engineering, University of Modena and Reggio Emilia, 41121 Modena, Italy.
        Email: {\tt\footnotesize \{mattia.catellani, lorenzo.sabattini\}@unimore.it}.}%
}

\begin{document}

\maketitle

\begin{abstract}
Control barrier functions (CBFs) are an effective model-based tool to formally certify the safety of a system. With the growing complexity of modern control problems, CBFs have received increasing attention in both optimization-based and learning-based control communities as a safety filter, owing to their provable guarantees. However,  success in transferring these guarantees to real-world systems is critically tied to model accuracy. For example, payloads or wind disturbances can significantly influence the dynamics of an aerial vehicle and invalidate the safety guarantee. In this work, we propose an efficient yet flexible online learning-enhanced high-order adaptive control barrier function using Neural ODEs. Our approach improves the safety of a CBF controller on the fly, even under complex time-varying model perturbations. In particular, we deploy our hybrid adaptive CBF controller on a $38\unit{g}$ nano quadrotor, keeping a safe distance from the obstacle, against $18\unit{km/h}$ wind.
\end{abstract}

\section{Introduction}
As the complexity of robotic mechanical systems and tasks increases, such as high-dimensional, nonlinear systems and agile control in quadrotors or quadruped robots, it leads to a broader adoption of learning-based approaches, which often come without safety guarantees. This necessitates formal safety guarantees built into the controller  by definition. 
Control barrier functions (CBFs) have become a popular model-based tool for safety-critical control applications owing to their guarantees on set invariance (i.e., safety) without compromising performance~\cite{ames2014control, ames2019control, wang2026connectivitymaintenancerecoverymultirobot}. That is, a safe control is computed by solving a quadratic program (QP) in a closed-loop system. 
In practice, however, the safety guarantee of a CBF can fail due to an inaccurate model. This inaccuracy is often caused by model mismatches or by external forces, like changes in payload or wind disturbances. 
The adaptive CBFs (aCBFs) compensate for model residuals and improve safety in real-world systems, addressing the aforementioned safety violations.

Inspired by data-driven approaches that learn complex models~\cite{o2022neural, chen2018neural, jiahao2021knowledge, jiahao2022learning}, we propose a learning-enhanced high-order adaptive CBF (HO-aCBF) that improves safety online under time-varying model perturbations. Our hybrid approach combines a nominal model with a neural network residual model to efficiently estimate the true dynamics using Knowledge-based Neural ODEs (KNODE)~\cite{jiahao2021knowledge}. The model trains on state-control data and allows data sampling with different time intervals (typical for a real-time system). Our approach offers high data efficiency compared to a pure learning-based approach by utilizing a nominal model, yet has the rich expressiveness of a neural network to capture complex residuals online. 

\newtext{
Neural-Fly~\cite{o2022neural} learns an adaptation term in its control law directly and achieves online adaptation to severe wind disturbances, but requires a significant offline training phase of approximately $12$-minute flight data under diverse wind conditions. In contrast, our method learns the model online and improves performance via a model-based controller, and does not require offline training, while supporting it when desired. Gaussian Processes (GPs) approaches, such as PILCO~\cite{deisenroth2011pilco}, learns dynamics for data-efficient policy search in reinforcement learning, and online GP residual learning~\cite{hewing2019cautious} has been applied for model predictive control. While the above methods provide alternative strategies for learning dynamics, they differ from our approach, which embeds learned dynamics directly into CBF constraints to improve safety online.
}
\begin{figure}[t]
    \centering
    \includegraphics[width=0.49\textwidth]{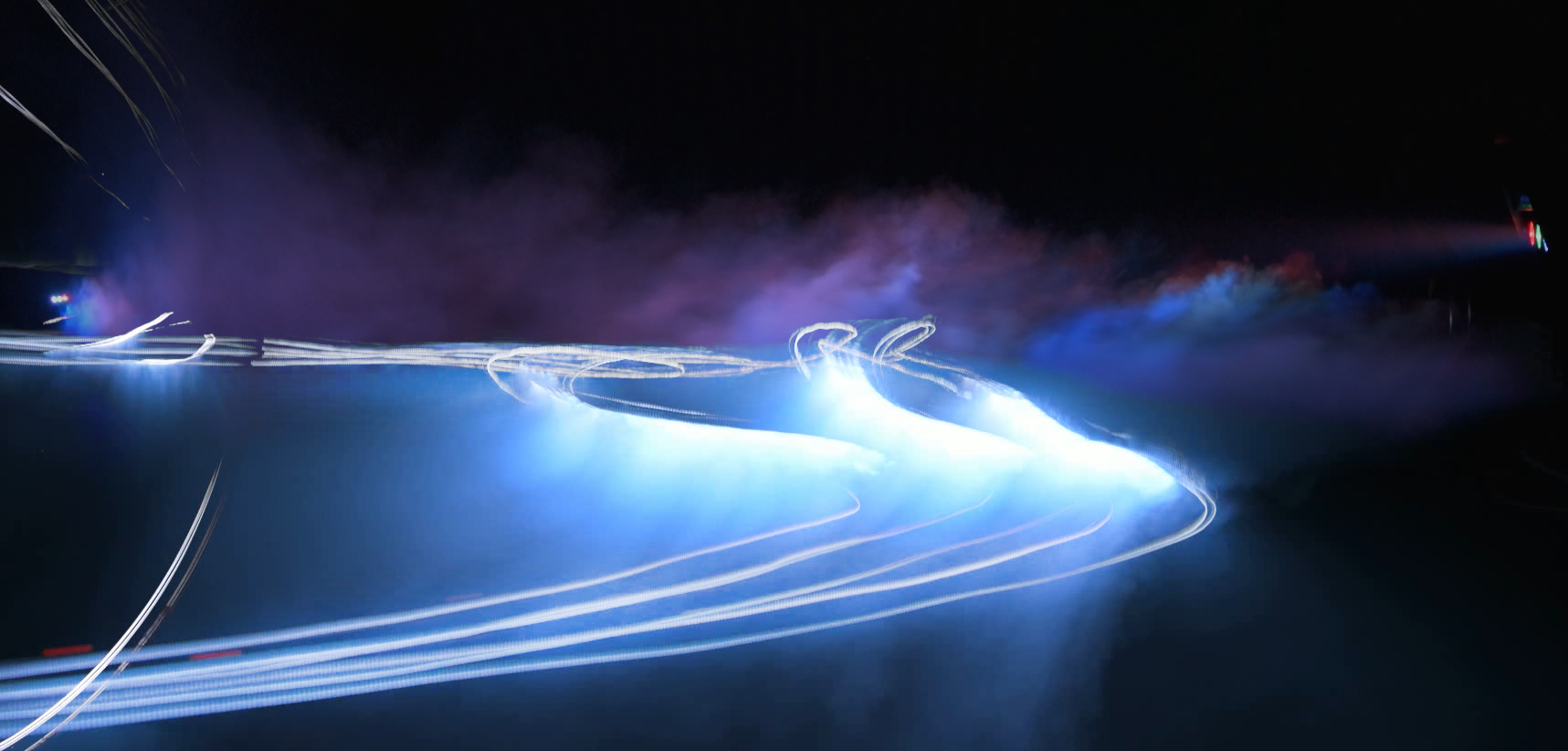}
    \caption{A 38$\unit{g}$ nano quadrotor tracks a circular trajectory while keeping a safe distance from the obstacle, against an $18\unit{km/h}$ wind. The safety is improved on-the-fly, i.e., the quadrotor moves further away from the obstacle after experiencing the wind once.}
    \label{fig:demo}
    \vspace{-1em}
\end{figure}
Recent studies have shown that aCBFs (specifically, residual learning approaches) are effective at improving safety under model perturbations~\cite{taylor2020learning,choi2020reinforcement}. However, these methods require offline training, which leads to safety violations when the residuals are out of the training distribution. 
Our learning-enhanced aCBF, owing to its hybrid model approach, can adapt to unseen residuals online, making it well-suited for real-world systems. 
In addition, in~\cite{taylor2020learning,choi2020reinforcement}, supervision of time derivatives of a control barrier function is required, which is obtained from noisy numerical differentiations. Instead, we use the Neural ODEs to approximate the residuals directly from observable state-control data. 
In practice, most CBF constraints for a mechanical system have relative degrees larger than one, thus requiring a high-order CBF (HOCBF). Our HO-aCBF algorithm generalizes the standard aCBF to handle high-order safety-critical constraints for a larger category of systems.

Robust CBFs guarantee safety against model perturbations by adding a robustifying term in CBF constraints~\cite{jankovic2018robust}. Robust CBFs, however, require knowing the bounds of the model uncertainty and  
result in overly conservative controllers. Recently, robust adaptive CBFs (RaCBFs) have been proposed~\cite{taylor2020adaptive, lopez2020robust, black2023safe}. In their formulation, an aCBF is used to mitigate over-conservativeness by estimating parametric uncertainty. 
The authors in~\cite{cohen2022high} extend the concurrent-learning aCBFs in~\cite{isaly2021adaptive} to HO-aCBFs. In these works, a strong assumption is made that the residual is a linear model combining a known knowledge matrix and a learnable constant vector. In practice, this matrix may not be available, and the uncertainty is not constant, thus breaking their safety guarantees. Extensive tuning is also required for the concurrent-learning approach. 

Unlike aCBF frameworks in~\cite{taylor2020adaptive, lopez2020robust, black2023safe, cohen2022high, isaly2021adaptive}, our approach does not require prior knowledge or assumption of residuals, captures time-varying dynamics, and is robust to tuning. 
In experiments, we demonstrate that HO-aCBF controllers are insufficient to capture time-varying residuals even with increased state-control data, resulting in inadequate performance. 
Additionally, our hybrid approach has high data efficiency compared to residual learning approaches in~\cite{taylor2020learning,choi2020reinforcement}, making it suitable for online adaptation. These learning frameworks~\cite{taylor2020learning,choi2020reinforcement} require estimating time derivatives of a control barrier function using numerical differentiation. Instead, 
our approach learns the residual dynamics directly from state-control data. 

The contributions of this work are twofold:
\begin{enumerate}
    \item a learning-enhanced high-order adaptive CBF that improves safety under complex time-varying model perturbations; and
    \item an online hybrid approach using Knowledge-based Neural ODEs that solves CBF optimization with learnable parameters at $100\unit{Hz}$ and trains online. 
\end{enumerate}
We demonstrate our controller's efficacy in simulation with different nonstationary residuals, benchmarked with HOCBFs~\cite{xiao2021high}, HO-aCBFs~\cite{isaly2021adaptive, cohen2022high} (with different hyperparameters). In physical experiments, we deploy our algorithm on a nano quadrotor weighing only $38\unit{g}$ against $18\unit{km/h}$ turbulent wind  and efficiently improve safety performance on the fly. 
\section{Preliminaries}
\subsection{High-order Control Barrier Functions (HOCBFs)}
Consider a nonlinear control affine system in the form
\begin{equation}\label{eq:nom_dynamics}
    \dot{\bf{x}} = f(\mathbf{x})+g(\mathbf{x})\mathbf{u},
\end{equation}
where $f: \mathbb{R}^p \rightarrow \mathbb{R}^p$ and $g: \mathbb{R}^p \rightarrow \mathbb{R}^{p\times q}$ are Lipschitz continuous functions, and $\mathbf{u} \in U \subset \mathbb{R}^q$ is the control input. 
A closed set $\mathcal{C}\in \mathbb{R}^{p}$ is \textit{forward invariant} for the closed-loop system $\dot{\mathbf{x}}$ if $\mathbf{x}(0)\in\mathcal{C}\Rightarrow \mathbf{x}(t)\in\mathcal{C}, \forall t$. In this paper, the \textit{forward invariance} property is used to formally define the notion of ``safety". It is assumed that any safe set can be expressed as a zero-superlevel set of a continuously differentiable function $h(\mathbf{x}): \mathbb{R}^p \rightarrow \mathbb{R}$, written as
\begin{align}
    \mathcal{C}=\{\mathbf{x}\in\mathbb{R}^{p} \mid h(\mathbf{x}) \geq 0\}. \label{eq:safe_set}
\end{align}

A control barrier function is a popular tool to produce controls that render the set~\eqref{eq:safe_set} forward invariant for~\eqref{eq:nom_dynamics}. Classical (order-one) CBF formulations, however, are conditioned on the assumption that $h$ has relative degree one with respect to~\eqref{eq:nom_dynamics}. 
\begin{definition}[Relative degree~\cite{khalil2002nonlinear}]\label{def:relative_degree} A sufficiently smooth function $h$ is said to have \textit{relative degree} $r\in \mathbb{N}$ with respect to~\eqref{eq:nom_dynamics} on the set $\mathbb{R}^{p}$ if 1) $L_{g}L_{f}^{i-1}h(\mathbf{x}) = 0$, for all $1\leq i \leq r-1$, and 2) $L_{g}L_{f}^{r-1}h(\mathbf{x}) \neq 0$ for all $\mathbf{x}\in \mathbb{R}^{p}$.
\end{definition}
For most mechanical systems and their safety requirements, $h$, a higher relative degree is required. High-order control barrier functions  provide safety for such systems. 
\begin{definition}[HOCBFs~\cite{xiao2021high, tan2021high}] \label{def:HOCBF} Consider a system as in~\eqref{eq:nom_dynamics}. Let $\{\mathcal{C}_{i}\}_{i=1}^{r}$ be a collection of sets of the form $\mathcal{C}_{i}=\{\mathbf{x}\in\mathbb{R}^{p}\mid \psi_{i-1}\geq0\}$ satisfying
\begin{align}
    \psi_{r}(\mathbf{x}) &= \dot{\psi}_{r-1}(\mathbf{x}) + \alpha_{r}(\psi_{r-1}(\mathbf{x}))\\
    \psi_{0}(\mathbf{x}) &= h(\mathbf{x}),
\end{align}
where $\{\alpha_{i}\}_{i=1}^{r}$ is a set of differentiable extended class $\mathcal{K}$ functions. A function $h$ is said to be a HOCBF of order $r$ for~\eqref{eq:nom_dynamics} on an open set $\mathcal{D} \supset \cap_{i=1}^{r}\mathcal{C}_{i}$ if $h$ has relative degree $r$ on some nonempty $\mathcal{R} \subseteq \mathcal{D}$ and there exists a suitable choice of $\{\alpha_{i}\}_{i=1}^{r}$, such that for all $\mathbf{x}\in \mathcal{D}$
\begin{align}
    \label{eq:hocbf_const}
    \sup_{\mathbf{u} \in U} [\underbrace{L_f \psi_{r-1}(\mathbf{x}) + L_g\psi_{r-1}(\mathbf{x})\mathbf{u} + \alpha_r(\psi_{r-1}(\mathbf{x}))}_{\psi_{r}(\mathbf{x}, \mathbf{u})}] \geq 0. 
\end{align}
\end{definition}
\begin{theorem}[\!\!\cite{tan2021high}] Let $h$ be a HOCBF for~\eqref{eq:nom_dynamics} on $\mathcal{D}\subset \mathbb{R}^{p}$ as in Def.~\ref{def:HOCBF}. Any locally Lipschitz controller $\mathbf{u} \in U$ that satisfies $\psi_{r}(\mathbf{x}, \mathbf{u})\geq 0$, renders $\cap_{i=1}^{r}\mathcal{C}_{i}$ \textit{forward invariant} for the closed-loop system as in~\eqref{eq:nom_dynamics}.
\end{theorem}
\subsection{Knowledge-based Neural ODEs (KNODE)}
KNODE~\cite{jiahao2021knowledge} is a data-driven approach that learns the model from state-control pairs, i.e., $\mathbf{z}:=[\mathbf{x};\mathbf{u}]$. Its data efficiency comes from leveraging a nominal model, learning merely the residual dynamics. The hybrid model is given by
\begin{align}\label{eq:hybrid_dynamics}
    \dot{\mathbf{x}} = \tilde{f}(\mathbf{z}) + \hat{f}_{\boldsymbol{\theta}}(\mathbf{z}) = f_h(\mathbf{z}, \hat{f}_{\boldsymbol{\theta}}(\mathbf{z})),
\end{align}
where $\tilde{f}$ denotes a nominal model and $\hat{f}_{\boldsymbol{\theta}}$ denotes the residual model estimated by a neural network parameterized with $\boldsymbol{\theta}$. The neural ODE-based approach learns the model of a dynamical system. Instead of learning in the state space, it learns its dynamical model directly and predicts the state using ODE solvers, thus allowing observations sampled with different time intervals (typical in a physical real-time system). An efficient learning approach preserves the physics of a dynamic model, making it well-suited for our online HO-aCBF controller. 
\section{Problem Formulation}
Consider a robot that follows a reference with a desired control $\mathbf{u}_{\mathrm{des}}$ and certifies the safety of the desired control using a CBF. The safety guarantee of a CBF can fail under unknown model perturbations. 
We adopt the formulation in~\cite{taylor2020adaptive}, and assume the true model as follows 
\begin{align}\label{eq:dynamics}
    \dot{\mathbf{x}} = f(\mathbf{x}) + g(\mathbf{x})\mathbf{u} + d(\mathbf{x}),
\end{align}
where $f(\mathbf{x}) + g(\mathbf{x})\mathbf{u}$ forms a nominal model and $d(\mathbf{x})$ includes all unmodeled residuals and is allowed to change over time. The problem we want to solve is to recover the safety guarantee of the CBF-based controller by identifying the model perturbation $d(\mathbf{x})$ using online state-control data.

We propose an efficient learning-enhanced high-order aCBF that compensates for unknown time-varying and nonlinear residuals $d(\mathbf{x})$ and improves the safety online. 
Our hybrid model combines the nominal with a neural network residual model $\hat{d}_{\boldsymbol{\theta}}$ parameterized with $\boldsymbol{\theta}$, using KNODE~\cite{jiahao2021knowledge}. The residual model is trained on observable state-control data, i.e., $\mathbf{z}:=[\mathbf{x};\mathbf{u}]$. Furthermore, we generalize the proposed aCBFs to high-order aCBFs to handle a larger category of systems. 
\begin{figure}[t]
    \centering
    {\includegraphics[width=0.45\textwidth]{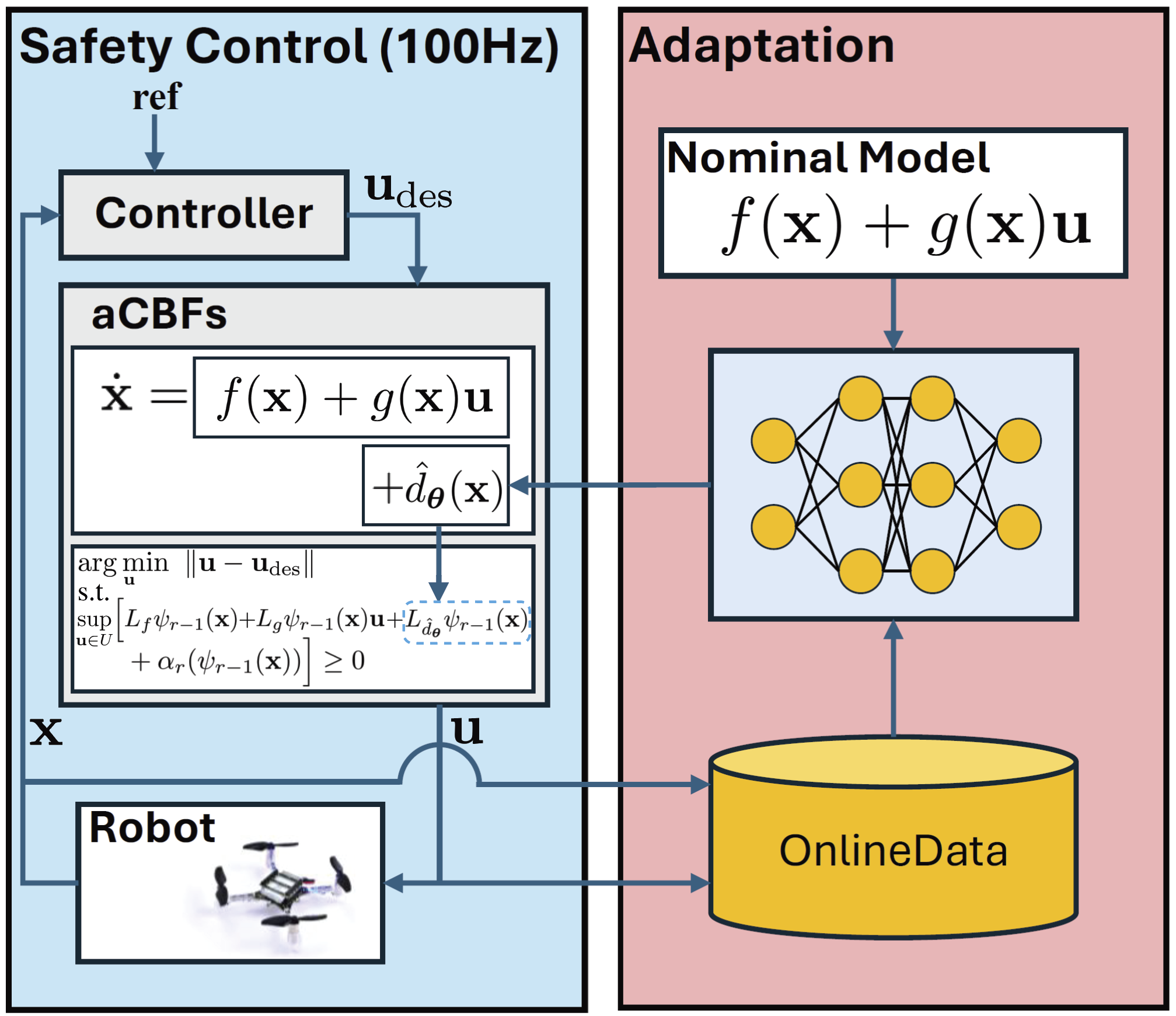}}
    \caption{The picture depicts the system overview of our NODE-HO-aCBF framework. The safety control solves the NODE-HO-aCBF QP, which uses the most recently trained model, at $100\unit{Hz}$ to certify the desired control. The state and control input data are stored in an online queue for training. The hybrid adaptation module combines a nominal model with a learning-based term that approximates the unknown residual dynamics online. 
    }
    \label{fig:framework}
    \vspace{-2em}
\end{figure}
\section{Learning-enhanced High-order Adaptive Control Barrier Functions}
Modeling inaccuracy can invalidate the safety guarantee of a CBF controller. 
We propose an efficient learning-enhanced high-order aCBF using KNODE, or NODE-HO-aCBF for short, depicted in Fig.~\ref{fig:framework}. Our approach compensates for model perturbations from state-control data, improving safety online. In Sec.~\ref{sec:Learning-enhanced_HO-aCBFs}, we introduce our NODE-HO-aCBF optimization framework. In Sec.~\ref{sec:KNODE_online}, we introduce KNODE for online residual learning. In Sec.~\ref{sec:method_example}, we present an example of collision avoidance using a double integrator with our controller. 
\subsection{Learning-enhanced HO-aCBFs}
\label{sec:Learning-enhanced_HO-aCBFs}
A simple nominal model is analytically convenient, but often does not capture the true dynamics. For instance, a quadrotor cannot exactly track a double integrator trajectory due to its kinematic constraints. 
External residual dynamics, such as wind turbulence or downwash effect between nearby quadrotors, can introduce nonlinear dynamics. 
Our approach allows an adaptive CBF to learn these model mismatches and external disturbances from data, improving safety online.  

We assume the true dynamics as in~\eqref{eq:dynamics}, where residual dynamics $d(\mathbf{x})$ depend on the state and are allowed to change over time. 
We define a hybrid model, combining the nominal model with a neural network residual model, as follows,
\begin{align}\label{eq:linear_hybrid_dynamics}
    f_h(\mathbf{z}, \hat{d}_{\boldsymbol{\theta}}(\mathbf{x})) = f(\mathbf{x}) + g(\mathbf{x})\mathbf{u} + \hat{d}_{\boldsymbol{\theta}}(\mathbf{x}),
\end{align}
where $\hat{d}_{\boldsymbol{\theta}}$ denotes the residual dynamics estimated by a neural network parameterized with $\boldsymbol{\theta}$. 

For a model-based controller, successfully transferring its guarantee to the real world depends on model fidelity. To handle model perturbations, we consider the hybrid model in~\eqref{eq:linear_hybrid_dynamics} and propose a Neural ODE enhanced high-order adaptive control barrier function, namely, NODE-HO-aCBF. Similar to an HOCBF QP, the objective is to minimize the deviation between the safe and the desired controls. To adapt to the unknown residual dynamics, we modify the HOCBF constraint in~\eqref{eq:hocbf_const} to~\eqref{eq:robust_cbf_constraints}. The Lie-derivative of residual term $\hat{d}_{\theta}$, i.e., $L_{\hat{d}_{\theta}}$, is added to the HOCBF constraint to certify the hybrid model. 
We formulate the NODE-HO-aCBF QP as follows,
\begin{subequations}\label{eq:NODE-HO-aCBF_optimization}
\begin{align}
    \argmin_{\mathbf{u}} ~ & \norm{ \mathbf{u} - \mathbf{u}_{\mathrm{des}} } \label{QPcost}\\
    \text{s.t.} 
    ~ & \sup_{\mathbf{u} \in U} \Big[L_{f} \psi_{r-1}(\mathbf{x}) + L_g\psi_{r-1}(\mathbf{x})\mathbf{u} + L_{\hat{d}_{\boldsymbol{\theta}}} \psi_{r-1}(\mathbf{x}) \nonumber \\ 
    & \quad \quad ~ + \alpha_{r}(\psi_{r-1}(\mathbf{x})) \Big] \geq 0,\ \label{eq:robust_cbf_constraints}
\end{align}
\end{subequations}
where,
\begin{align}
    \psi_{r} & = \dot{\psi}_{r-1} + \alpha_{r}(\psi_{r-1})\\
    & = \left(\frac{\partial \psi_{r-1}}{\partial \mathbf{x}}\right)^{\top}(f+g\mathbf{u}+\hat{d}_{\boldsymbol{\theta}}) + \alpha_{r}(\psi_{r-1})\\
    \psi_{0} & = h(\mathbf{x}).
\end{align}
To compute $\psi_{r}$, the algorithm requires taking the $(r\!-\!1)$-th derivative of the residual dynamics $\hat{d}_{\boldsymbol{\theta}}$ estimated by the neural network w.r.t. the state, i.e. $\nabla_{\mathbf{x}}^{r-1} \hat{d}_{\boldsymbol{\theta}}$, which can be computed by automatic differentiation from machine learning toolboxes or numerical methods. Our learning-enhanced controller runs at $100\unit{Hz}$ on a real-time system. At each iteration step (see Algo.~\ref{alg:data_and_model_update}), the NODE-HO-aCBF loads the hybrid model if a new learned model is ready. A PID controller takes the current state $\mathbf{x}$ and reference $\mathbf{x}_{\mathrm{ref}}$ to generate a desired control $\mathbf{u}_{\mathrm{des}}$. By solving the QP in~(\ref{eq:NODE-HO-aCBF_optimization}), a safe control $\mathbf{u}$ is produced and executed by the robot. The state-control pair $\mathbf{z}:=[\mathbf{x}, \mathbf{u}]$ is added to the online queue for training. 
\begin{algorithm}[t]
\caption{NODE-HO-aCBF Control Loop}
\label{alg:data_and_model_update}
Initialize the current time, latest model update time, total duration and training data as $t_{i}$, $t_{m}$, $t_{T}$, and $\mathcal{Q}_{\mathrm{train}}$\\
$t_{i} \leftarrow 0$; $\mathcal{Q}_{\mathrm{train}} \leftarrow [~]$\\
\While{$t_{i} < t_{T}$} {
    \If {New hybrid model is available} {
        NODE-HO-aCBF loads the new model
        }
    PID generate a desired control $\mathbf{u}_{\mathrm{des}}$ given robot state $\mathbf{x}$ and a reference $\mathbf{x}_{\mathrm{ref}}$\\
    Solve QP in~(\ref{eq:NODE-HO-aCBF_optimization}) and generate a safe control $\mathbf{u}$\\
    Robot updates the state using $\mathbf{u}$\\
    Append $\mathbf{z}_{t_{i}}$ to the queue $\mathcal{Q}_{\mathrm{train}}$ (pop out the oldest data if the queue is full).\\
    $t_{i}\leftarrow$ current time
}
\end{algorithm}
\begin{algorithm}[t]
\caption{Online Dynamic Learning}
\label{alg:online_learning_model}
Initialize the current time, total duration, training data as $t_{i}$, $t_{T}$, $\mathcal{Q}_{\mathrm{train}}$, and the hybrid model in~\eqref{eq:linear_hybrid_dynamics}\\
\While{$t_{i} < t_{T}$} {
    \If{$\mathcal{Q}_{\mathrm{train}} \neq \emptyset$} {
        Train the hybrid model by optimizing~\eqref{eq:NODE_optimization} using $\mathcal{Q}_{\mathrm{train}}$\\
        Save the new model
    }
    $t_{i}\leftarrow$ current time
}
\end{algorithm}
\subsection{KNODE Online Learning}
\label{sec:KNODE_online}
An unknown nonstationary residual, as prior models quickly become obsolete, requires continuous online adaptation to identify the current true dynamics. 
We employ a queue $\mathcal{Q}_{\mathrm{train}}$ to store state-control pairs in a first-in, first-out (FIFO) manner, ensuring that the hybrid model $f_{h}$ is trained online with the most recent data (see Algo.~\ref{alg:online_learning_model}). 
The neural ODE learns the dynamical system directly. Consider an objective function $\mathcal{L}$ that penalizes the error between the predictions and observations, where the prediction obeys a learning-based ODE, defined as follows, 
\begin{align}\label{eq:loss}
    \mathcal{L}(\boldsymbol{\theta})=&\frac{1}{m-1} \sum_{i=1}^{m-1} \int_{t_i}^{t_{i+1}} \delta\left(t_s-\tau\right)\|\hat{\mathbf{x}}_{\tau}-\mathbf{x}_{\tau}\|^2 d \tau + \lambda \norm{\boldsymbol{\theta}}_{2}^{2}
\end{align}
where $m$ is the number of samples in $\mathcal{Q}_{\mathrm{train}}$, $\delta(\cdot)$ is the Dirac delta function, $t_{s}$ is any sampling time in $\mathcal{Q}_{\mathrm{train}}$, $\mathbf{x}_{\tau}$ is a shorthand for $\mathbf{x}(\tau)$. For the learned model to generalize, we use the L2 regularization with a weight $\lambda$~\cite{zhang2016understanding}. The prediction is computed as follows, 
\begin{align}\label{eq:prediction}
    \hat{\mathbf{x}}_{\tau} = \mathbf{x}_{t_{i}} + \int_{t_{i}}^{\tau} f_h(\mathbf{z}_{w}, \hat{d}_{\boldsymbol{\theta}}(\mathbf{x}_{w}))dw.
\end{align}
The parameters $\boldsymbol{\theta}$ are obtained by solving the optimization:
\begin{subequations}\label{eq:NODE_optimization}
\begin{align}
    \argmin_{\boldsymbol{\theta}} ~ & \mathcal{L}(\boldsymbol{\theta}) \label{eq:NODE_loss}\\
    \text{s.t.} 
    ~ &  \dot{\mathbf{x}} = f_{h}(\mathbf{z}, \hat{d}_{\boldsymbol{\theta}}(\mathbf{x})) .\label{eq:NODE_const}
\end{align}
\end{subequations}
In practice, a gradient-based optimization tool, such as PyTorch, can be employed to solve the above optimization.

Our algorithm supports offline pre-training. In offline mode, we collect state-control data by using the HOCBF safety controller. We pre-train the neural network to approximate the residual dynamics using all the data collected and deploy the NODE-HO-aCBF controller with the learned $\hat{d}_{\boldsymbol{\theta}}(\mathbf{x})$.
\section{Simulations}
To demonstrate the efficacy of our algorithm, we design a simulation with a double integrator with different residuals. 
\subsection{NODE-HO-CBFs with Double Integrator Nominal Model}
\label{sec:method_example}
A robot is required to follow a reference trajectory in a workspace containing obstacles $\mathcal{O}_{1},\ldots, \mathcal{O}_{N_{\mathrm{obs}}}$, from which the robot is required to keep a safety distance $D_s$.
The state $\mathbf{x} = [\mathbf{r}; \dot{\mathbf{r}}]\in \mathbb{R}^{6}$ contains position and velocity. The control $\mathbf{u}\in \mathbb{R}^{3}$ is the acceleration. We consider the nominal model as a double integrator, and the hybrid model can be written as
\begin{align}\label{eq:DI_hybrid}
    \dot{\mathbf{x}} &= A\mathbf{x} + B\mathbf{u} + \hat{d}_{\boldsymbol{\theta}}(\mathbf{x}),
\end{align}
where $A = [\mathbf{0}, \mathbf{I}; \mathbf{0}, \mathbf{0}] \in \mathbb{R}^{6\times 6}$, $B = [\mathbf{0}; \mathbf{I}] \in \mathbb{R}^{6\times 3}$. 
Here $\mathbf{0} \in \mathbb{R}^{3\times 3}$, $\mathbf{I} \in \mathbb{R}^{3\times 3}$ are the zero and identity matrices, respectively. 
\newtext{We utilize this simplified model to demonstrate how the residual learning term $\hat{d}_{\boldsymbol{\theta}}(\mathbf{x})$ effectively captures unmodeled nonlinearities and disturbances, ensuring safety even with inaccurate nominal dynamics.}

At each step, we obtain reference position, velocity, and acceleration, i.e., $[\mathbf{r}_{\mathrm{des}}; \mathbf{v}_{\mathrm{des}}; \mathbf{a}_{\mathrm{des}}]$. 
The desired control $\mathbf{u}_{\mathrm{des}}$ is generated using PID. We investigate the CBF that ensures safety by maintaining a minimum distance from obstacles. 
Specifically, we define a CBF for the $i$-th obstacle as
\begin{align}
    h_{i}(\mathbf{x}) = \norm{\mathbf{r}-\mathbf{r}_{\mathrm{obs},i}}_{2}^{2} - D_{s}^{2}, \forall i\in\{1,\ldots N_{obs}\},
\end{align}
where $\mathbf{r}_{\mathrm{obs},i}$ denotes the position of $i$-th obstacle\newtext{, and $D_{s}$ is the minimal safety distance between the robot and obstacles}. The relative degree of $h_{i}$ is two with respect to the model in~\eqref{eq:DI_hybrid}. We consider the extended class $\mathcal{K}$ functions as $\alpha_{i}(h) = \gamma_{i}h$. The optimization is formulated as follows:
\begin{subequations}\label{eq:optimization}
\begin{align}
    \argmin_{\mathbf{u}} ~ & \norm{ \mathbf{u} - \mathbf{u}_{\mathrm{des}} } \label{QPcost_DI}\\
    \text{s.t.} 
    ~ & \sup_{\mathbf{u} \in U} \Big[(L_{f}^{2}h+L_{\hat{d}_{\boldsymbol{\theta}}}^{2}h) +L_{f}L_{\hat{d}_{\boldsymbol{\theta}}}h+L_{\hat{d}_{\boldsymbol{\theta}}}L_{f}h \nonumber\\
    &  ~+ [L_{g}(L_{f}h + L_{\hat{d}_{\boldsymbol{\theta}}}h)]\mathbf{u} 
    +(\gamma_{1}+\gamma_{2})(L_{f}h+L_{\hat{d}_{\boldsymbol{\theta}}}h)\nonumber\\
    &  ~ +\gamma_{2}\gamma_{1}h \Big] \geq 0.
\end{align}
\end{subequations}
The solution of the optimization \eqref{eq:optimization} is the control $\mathbf{u}$ for our hybrid learning-enhanced high-order aCBF controller.

\subsection{Simulation Setups}
A perturbed double integrator is implemented as $\dot{\mathbf{x}} = A\mathbf{x}+B\mathbf{u}+d(\mathbf{x})$, where we assume the residual $d(\mathbf{x})$ is unknown to the robot in the simulation. Three types of residuals are considered, namely, ``Attractive", ``Repulsive", and ``Time-varying". In ``Attractive" and ``Repulsive" experiments, the robot is attracted and repulsed towards and away from the center of a \textcolor{black}{forbidden region where the robot is not allowed to enter}, i.e., $d_{\mathrm{att}}(\mathbf{x})=[\mathbf{0}; -k \mathbf{r}]$, $d_{\mathrm{rep}}(\mathbf{x})=[\mathbf{0}; k \mathbf{r}]$, where $k\!=\!0.4$. In ``Time-varying" experiment, residuals change between repulsive and attractive over time, i.e., $d_{\mathrm{var}}(\mathbf{x})=[\mathbf{0}; k\sin(t)\mathbf{r}]$, where $k\!=\!1.0$. 
A Runge–Kutta (RK4) numerical solver with sampling time of  $\Delta t=0.01\unit{s}$ is used to simulate responses given the control. 
We assume  the robot state-control pair $\mathbf{z}:=[\mathbf{x};\mathbf{u}]$ can be measured at $100\unit{Hz}$. The closed-loop control runs at $100\unit{Hz}$. We use a similar neural network architecture to~\cite{jiahao2023online}. The value of $\hat{d}_{\boldsymbol{\theta}}$ is computed as the \newtext{weighted sum} of three parallel neural networks with the same architecture. Each neural network has two hidden layers, each with only $16$ neurons, where the hyperbolic tangent activation is applied after the first hidden layer. \newtext{The parallel structure enables continual adaptation while preserving previously learned dynamics. The queue size is fixed to three: in each training epoch, a new neural network is added, and the oldest one is removed if the queue is at capacity. Older neural networks are combined using exponential forgetting weights. This facilitates online adaptation and avoids instability in the closed-loop controller.}

\newtext{A closed-loop PID synthesizes the desired control $\mathbf{u}_{\mathrm{des}}$ to track a circular reference of radius $2\unit{m}$ at height $1\unit{m}$, centered at the origin of the x-y plane, with constant angular velocity $0.5\unit{rad/s}$. A spherical forbidden region of radius $D_s = 3\unit{m}$ is also located at the origin. }
Thus, the CBF must be effective to avoid safety violations. 
Our NODE-HO-aCBF controller supports both offline and online modes as described in Sec.~\ref{sec:KNODE_online}. For our online controller, we maintain an FIFO queue of $10\unit{s}$ of online data for training. For our offline model, we collect $40\unit{s}$ of state-control data for training.

\noindent\textbf{Benchmarking.} We compare the safety performance between: 1) HOCBFs baseline without residuals~\cite{xiao2021high} (we use the open-source implementation from~\cite{pan2025robust}), 2) HOCBFs baseline, 3) HO-aCBFs baseline~\cite{isaly2021adaptive, cohen2022high}, 4) our online NODE-HO-aCBFs, and 5) our offline NODE-HO-aCBFs. 
The HOCBF controller without external residuals acts as the baseline performance. 
The comparison of our online and offline controllers with HOCBFs baseline demonstrates that our controllers improve safety by adapting to residuals. The comparison of our online and offline controllers with adaptive HOCBFs highlights that our controllers can handle time-varying residuals, without knowledge of residuals nor extensive tuning. 
In CBF constraints of all controllers, we use extended class $\mathcal{K}$ functions $\alpha_{i}(h) = \gamma_{i}h$, where $\gamma_{i}=2$ for $i=1,2$. For the HO-aCBFs baseline~\cite{isaly2021adaptive, cohen2022high}, it is assumed that $d(\mathbf{x})=Y(\mathbf{x})\hat{\boldsymbol{\vartheta}}$, where $Y(\mathbf{x})\in \mathbb{R}^{p \times s}$ is a known matrix and $\hat{\boldsymbol{\vartheta}}\in \mathbb{R}^{s}$ is a constant vector to estimate. To update the parameter $\hat{\boldsymbol{\vartheta}}$, HO-aCBFs require computing its update law $\dot{\hat{\boldsymbol{\vartheta}}}$ using an adaptation gain $\kappa \in \mathbb{R}_{>0}$ and a positive definite gain matrix $\Gamma \in \mathbb{R}^{s \times s}$ (see~\cite[eq.~(14)]{cohen2022high}). We extensively tune $\kappa$ and $\Gamma$ in $\dot{\hat{\boldsymbol{\vartheta}}}$ and assume a known $Y(\mathbf{x})=[\mathbf{0}_{3\times 3};-r_{1},0,0;0,-r_{2},0;0,0,-(r_{3}-1)]$. Note that this is a strong assumption and we usually do not have prior knowledge about $Y(\mathbf{x})$ in practice. To compute the HO-aCBF controller online at $100\unit{Hz}$, we can only use a history stack of $0.5\unit{s}$. In the following section, we discuss that increasing the HO-aCBF's history stack does not improve its performance significantly, yet prohibiting its online performance. 
For HO-aCBF baseline~\cite{cohen2022high}, we set the bounds of the estimate $\hat{\boldsymbol{\vartheta}}$ between $[-4,-4,-4]$ to $[4,4,4]$, which cover the range of the underlying residual given our choice of $Y(\mathbf{x})$. 
\begin{table*}[!t]
\centering
\scalebox{0.85}{
\begin{tabular}{|c||c|c|c|c|c||c|c|c|c|c||c|c|c|c|c|}
\hline
\multicolumn{1}{|c||}{Residual Type}  & \multicolumn{5}{c||}{ Attractive } & \multicolumn{5}{c||}{ Repulsive } & \multicolumn{5}{c|}{ Time-varying }\\ 
\hline
\multicolumn{1}{|c||}{Method} & \multicolumn{1}{c|}{$h_{\mathrm{min}}$} & \multicolumn{1}{c|}{$h_{\mathrm{neg}}$} & \multicolumn{1}{c|}{Avg.} & \multicolumn{1}{c|}{Avg.} & \multicolumn{1}{c||}{S.sDist.}  & \multicolumn{1}{c|}{$h_{\mathrm{min}}$} & \multicolumn{1}{c|}{$h_{\mathrm{neg}}$} & \multicolumn{1}{c|}{Avg.} & \multicolumn{1}{c|}{Avg.} & \multicolumn{1}{c||}{S.sDist.} & \multicolumn{1}{c|}{$h_{\mathrm{min}}$} & \multicolumn{1}{c|}{$h_{\mathrm{neg}}$} & \multicolumn{1}{c|}{Avg.} & \multicolumn{1}{c|}{Avg.} & \multicolumn{1}{c|}{S.sDist.}  \\
\multicolumn{1}{|c||}{} & \multicolumn{1}{c|}{} & \multicolumn{1}{c|}{} & \multicolumn{1}{c|}{Dist.} & \multicolumn{1}{c|}{sDist.} & \multicolumn{1}{c||}{Var.}  & \multicolumn{1}{c|}{} & \multicolumn{1}{c|}{} & \multicolumn{1}{c|}{Dist.} & \multicolumn{1}{c|}{sDist.} & \multicolumn{1}{c||}{Var.} & \multicolumn{1}{c|}{} & \multicolumn{1}{c|}{} & \multicolumn{1}{c|}{Dist.} & \multicolumn{1}{c|}{sDist.} & \multicolumn{1}{c|}{Var.}  \\
\hline
CBFs No Res. & 0.09 & 0.0\% & 0.07 & 0.01 & 2.02e-10
& 0.09 & 0.0\% & 0.07 & 0.01 & 2.02e-10
& 0.09 & 0.0\% & 0.07 & 0.01 & 2.02e-10\\
\hline
CBFs &  -1.42 & 95.8\% & 0.27 & 0.25 & 8.20e-11 
& 2.34 & 0.0\% & 0.42 & 0.37 & 5.64e-10
& -2.78 & 44.7\% & 0.45 & 0.39 & 0.18 \\
\hline
aCBFs (Tuned) & -0.12 & 93.0\% & \textbf{0.06} & \textbf{0.01} & 9.89e-6 
& -0.10 & 89.5\% & \textbf{0.07} & \textbf{0.01} & 9.89e-6 
& -1.98 & 49.7\% & 0.27 & 0.23 & 0.06 \\
\hline
aCBFs $\Gamma=1$ & -1.50 & 92.6\% & 0.19 & 0.15 & 1.63e-3 
& 13.31 & 0.0\% & 1.49e+9 & 2.97e+9 & 4.69e+19 
& -6.98 & 0.0\% & 1.47e+6 & 2.94e+6 & 3.74e+13 \\
\hline
aCBFs Const. $Y$ & -0.93 & 95.1\% & 0.19 & 0.16 & 1.44e-6 
& 0.84 & 0.0\% & 0.24 & 0.18 & 5.29e-6  
& -2.65 & 45.4\% & 0.41 & 0.36 & 0.16 \\
\hline
Ours Online & -0.87 & 46.8\% & \textbf{0.07} & \textbf{0.02} & 3.93e-4 
& -1.61 & 37.7\% & 0.13 & \textbf{0.03} & 9.74e-4
& -2.98 & 52.3\% & 0.20 & \textbf{0.04} & \textbf{2.29e-3} \\
\hline
Ours Offline & \textbf{0.01} & \textbf{0.0\%} & \textbf{0.06} & \textbf{0.01} & 1.45e-5 
& \textbf{0.01} & \textbf{0.0\%} & \textbf{0.06} & \textbf{0.01} & 1.62e-5
& \textbf{-0.91} & 68.5\% & \textbf{0.10} & \textbf{0.05} & 3.11e-3 \\
\hline
\end{tabular}
}
\caption{Quantitative comparison between the baseline controllers and our controllers \newtext{(The ``HO-" prefix is dropped in all methods due to space limit)} with different residuals in simulation. The statistics are averaged across 10 trials.}
\label{table:quantitative}
\vspace{-2em}
\end{table*}
\subsection{Simulation Qualitative Results} 
\label{sec:sim_qual_res}
\begin{figure}[tb]
    \centering
    {\includegraphics[width=0.48\textwidth]{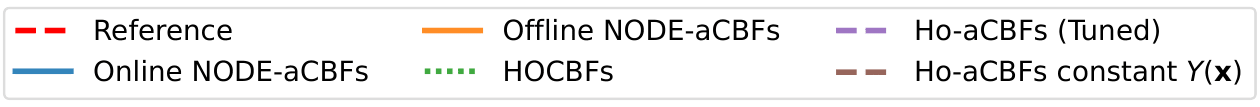}}\\
    {\includegraphics[width=0.244\textwidth]{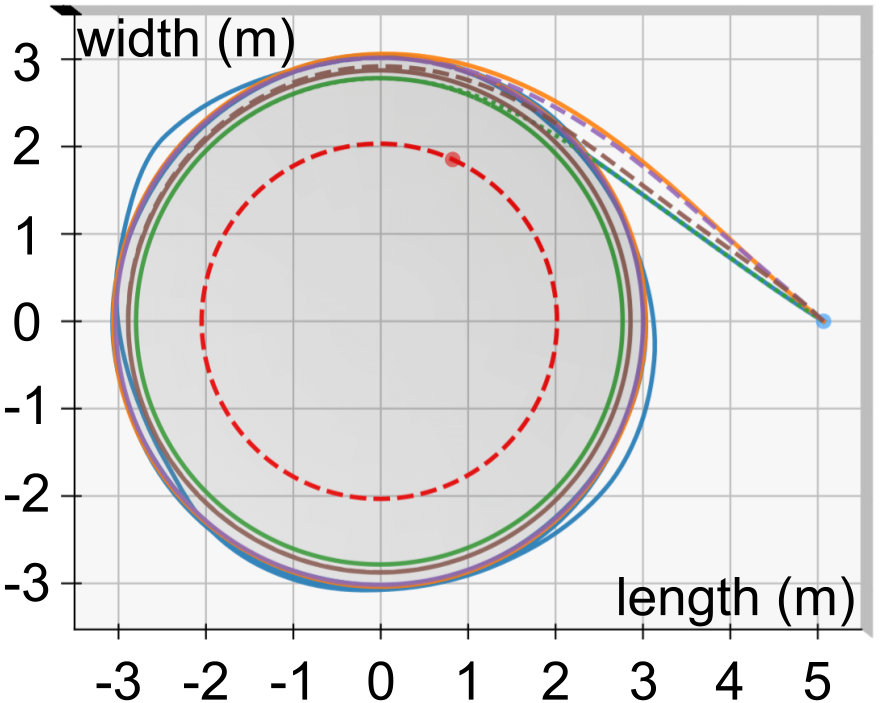}}
    {\includegraphics[width=0.232\textwidth]{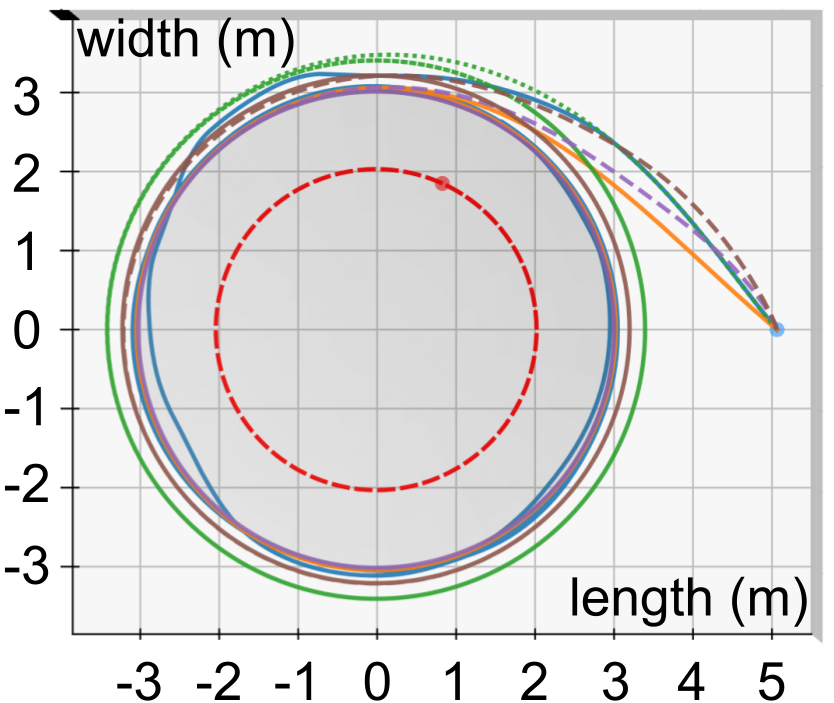}}\\
    {\includegraphics[width=0.261\textwidth]{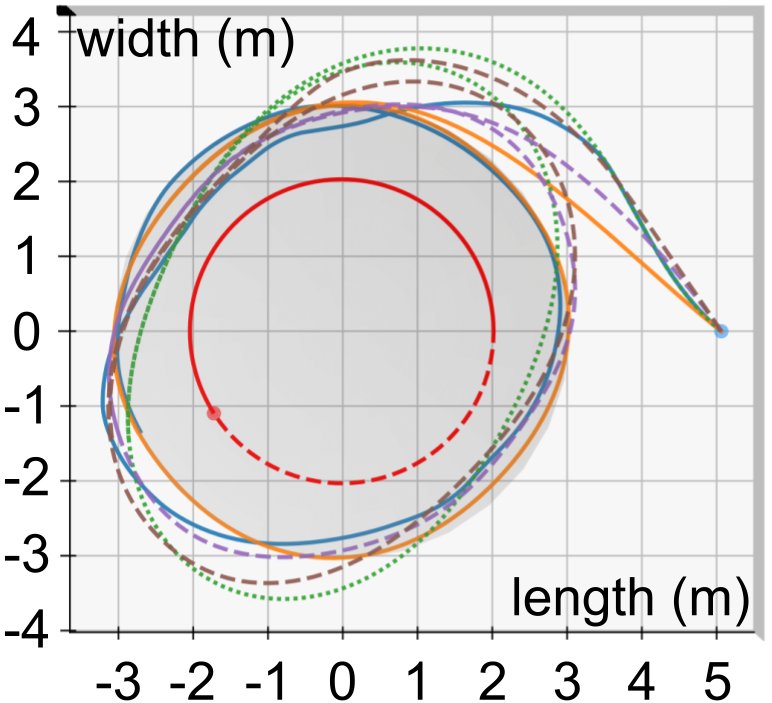}}
    {\includegraphics[width=0.215\textwidth]{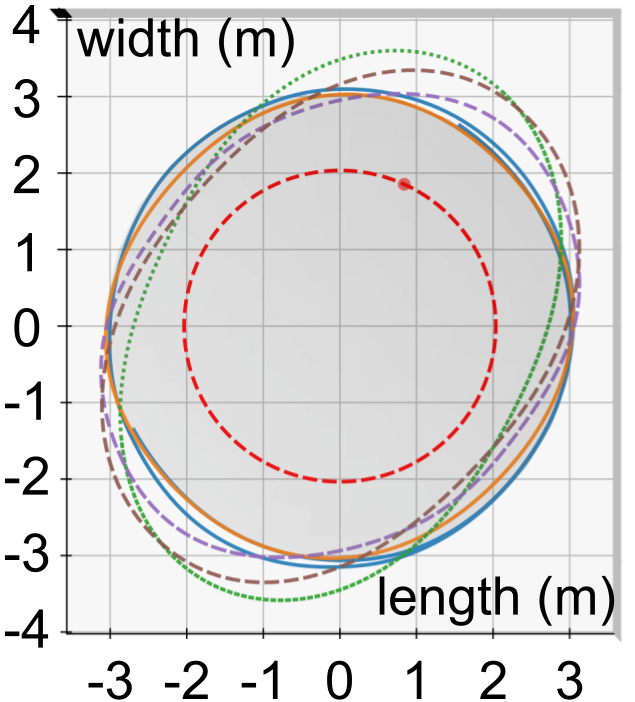}}
    \caption{Trajectories of safety controllers with different residuals. The blue dot represents the initial state, and the red dot represents the current reference. The gray sphere is the forbidden region. Top left: ``Attractive" residual, Top right: ``Repulsive" residual, Bottom left: ``Time-varying" residual from $0$-$20\unit{s}$, Bottom right: ```Time-varying" residual from $20$-$40\unit{s}$.}
    \label{fig:qualitative}
    \vspace{-1em}
\end{figure}
\begin{figure}[tb]
    \centering
    {\includegraphics[width=0.49\textwidth]{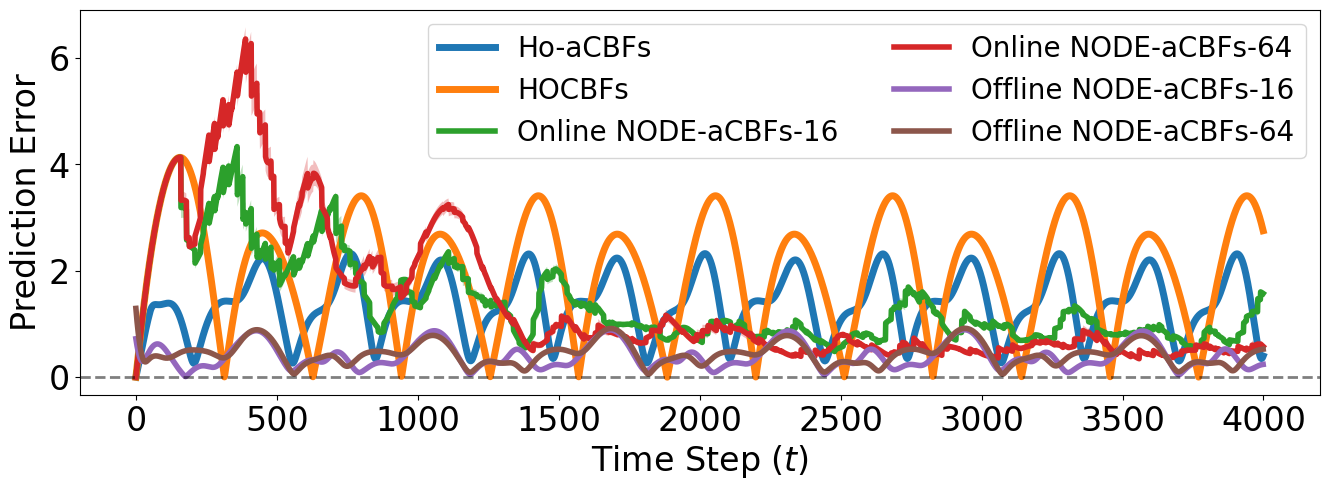}}
    \caption{\newtext{Time series of ``Time-varying" residual prediction error of different controllers. The number attached to our controller stands for the number of neurons in each hidden layer. The 95\% confidence intervals are over $10$ trials.}}
    \label{fig:rebuttal_time_series}
    \vspace{-2em}
\end{figure}
We demonstrate the typical trajectories of baseline and our NODE-HO-aCBF controllers with three residual types in Fig.~\ref{fig:qualitative}. 
In ``Attractive" experiments (top-left), the HOCBF baseline fails to adapt, resulting in safety violations, as shown by the green trajectory. The HO-aCBFs, when well-tuned and given a high-fidelity $Y(\mathbf{x})$, can compensate for residuals. 
With a constant $Y(\mathbf{x})$, the HO-aCBF controller violates safety, showcasing reduced adaptability, as indicated by the brown trajectory. Both our online and offline controllers adapt to residuals and settle outside of the forbidden region without tuning. In ``Repulsive" experiments (top-right), instead of being attracted to the center and violating safety, the HOCBF controller settles farther from the center. The HO-aCBF controller can only adapt with a high-fidelity $Y(\mathbf{x})$ and well-tuned. Our controllers, again, can adapt and settle at the boundary. \newtext{Note that the NODE-HO-aCBF controller results in a noiser and slower adaptation at the beginning, as in Fig.~\ref{fig:qualitative}, that the early trajectory can violate safety in the ``Repulsive" experiment. The HO-aCBFs update $\hat{\boldsymbol{\vartheta}}$ in every control loop, instead, NODE-HO-aCBFs learn a neural network alongside the controller. We call this noisier and slower convergence caused by a larger parameter space and alongside learning as learning delay. Despite this delay, the neural network provides nonlinearity required by complex residuals and learns an underlying manifold, so that the knowledge can be generalized to unseen areas and can learn patterns from more data without slowing down the control loop.}
We highlight that our controllers successfully adapt to ``Time-varying" residual and have significant advantages in settling closer to the safety boundary, i.e., improve the safety, while all the other controllers fail to adapt. 
\newtext{To highlight the performance of our controller. We plot the prediction error defined as $\ell_{2}$-norm of the difference between $d(\mathbf{x})$ and $\hat{d}(\mathbf{x})$ in ``Time-varying" experiments. Our online and offline controllers achieve lower prediction errors compared to the HOCBF and HO-aCBF controllers. Increasing the number of neurons in our controller adapts better with more data in ``Time-varying" experiments, but with a noisier adaptation at the start.}

\begin{figure}[t]
    \centering
    {\includegraphics[width=0.255\textwidth]{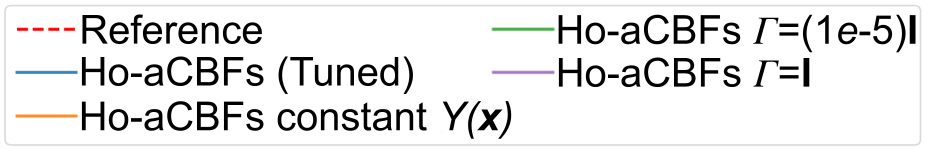}}
    {\includegraphics[width=0.223\textwidth]{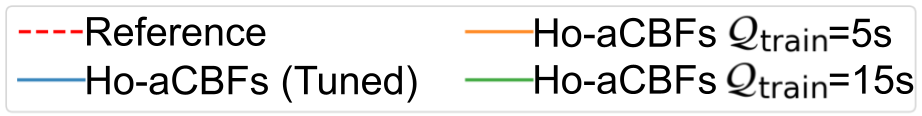}}\\
    {\includegraphics[width=0.256\textwidth]{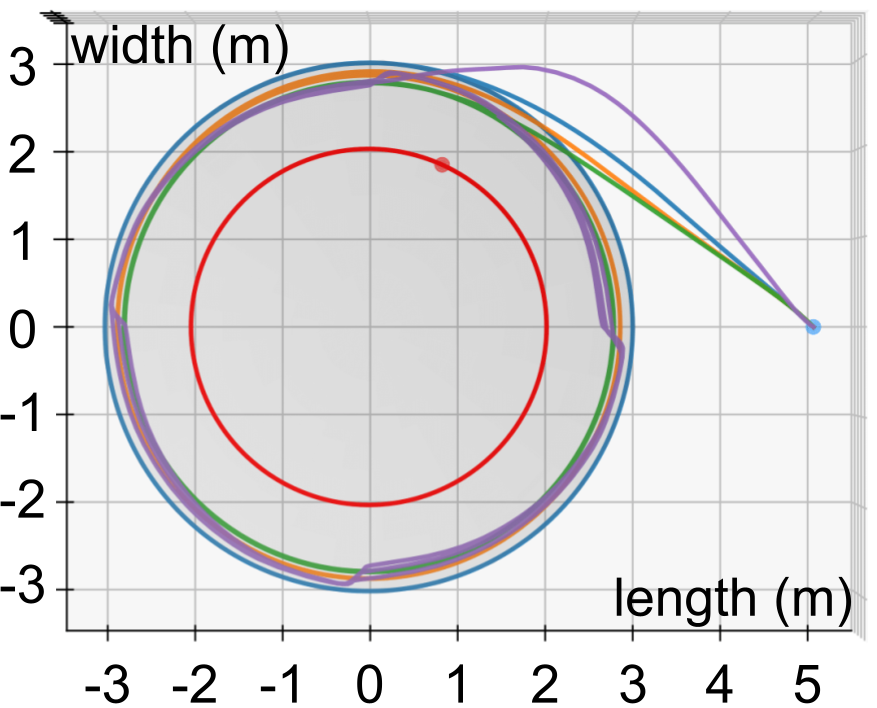}}
    {\includegraphics[width=0.22\textwidth]{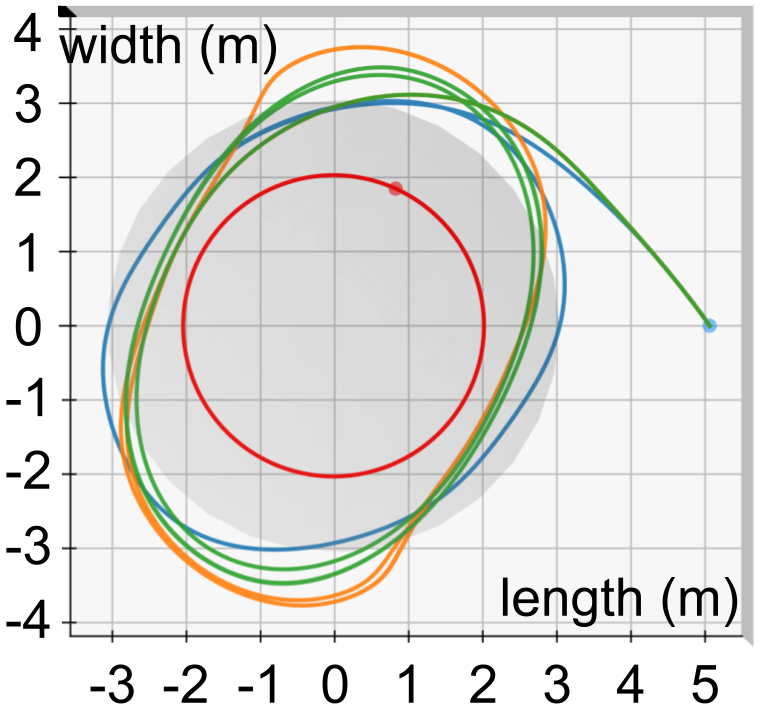}}
    \caption{Qualitative comparison between baseline HO-aCBF controller with different settings. ``Attractive" residual (on the left), and ``Time-varying" residual (on the right) are applied in experiments.}
    \label{fig:qualitative_hoacbf_analysis}
    \vspace{-2em}
\end{figure}

Next, we discuss tuning fragility, strong dependence on assumptions, and limitation on expressiveness and allowed data size of the HO-aCBF baseline. 
We tune $\kappa$ and $\Gamma$, and provide a strong assumption of the knowledge matrix $Y(\mathbf{x})$. 
Fig.~\ref{fig:qualitative_hoacbf_analysis} (left) shows controllers' performance in the ``Attractive" experiment. Note that a small $\Gamma=(1e\!-\!5)\mathbf{I}$, a large $\Gamma=\mathbf{I}$, or a constant matrix $Y(\mathbf{x})=[\mathbf{0}_{3\times 3};-\mathbf{I}_{3\times 3}]$ leads to a degenerated performance of HO-aCBFs. In Fig.~\ref{fig:qualitative_hoacbf_analysis} (right), we increase the history stack of a tuned HO-aCBF controller to $5\unit{s}$ and $15\unit{s}$ (a large history stack \newtext{significantly slows down the control step}, so we freeze the simulation when computing the HO-aCBF control). \newtext{We show the controller runtime and neural network training time in Table~\ref{table:runtime}.} However, the controller cannot utilize more data and safety violations preserved. These prominent drawbacks---fragility to tuning,  strong assumption, and limited expressiveness in pattern recognition and allowed history window---make it secondary compared to our approach. 
\begin{table}[h]
\centering
\scalebox{0.85}{
\begin{tabular}{|c|c|c|c|}
\hline
\multicolumn{1}{|c|}{Methods} & \multicolumn{1}{c|}{Controller runtime ($\unit{ms}$)} & \multicolumn{1}{c|}{Training time ($\unit{s}$)} & \multicolumn{1}{c|}{Data size ($\unit{s}$)} \\
\hline
Ours Online - $16$ & $\textbf{1.36}$ & $0.55$ & $10.00$ \\
\hline
Ours Online - $64$ & $2.35$ & $1.33$ & $10.00$ \\
\hline
HO-aCBF & 9.93 & - & $0.50$ \\
\hline
HO-aCBF & 32.40  & - & $5.00$ \\
\hline
HO-aCBF & 50.81  & - & $10.00$ \\
\hline
\end{tabular}
}
\caption{\newtext{Controller and training runtimes for different algorithms. The number attached to our controller stands for the number of neurons in each hidden layer. The statistics are averaged over $3000$ control steps.}}
\label{table:runtime}
\vspace{-2em}
\end{table}
\begin{figure*}[t]
    \centering
    {\includegraphics[width=0.161\textwidth]{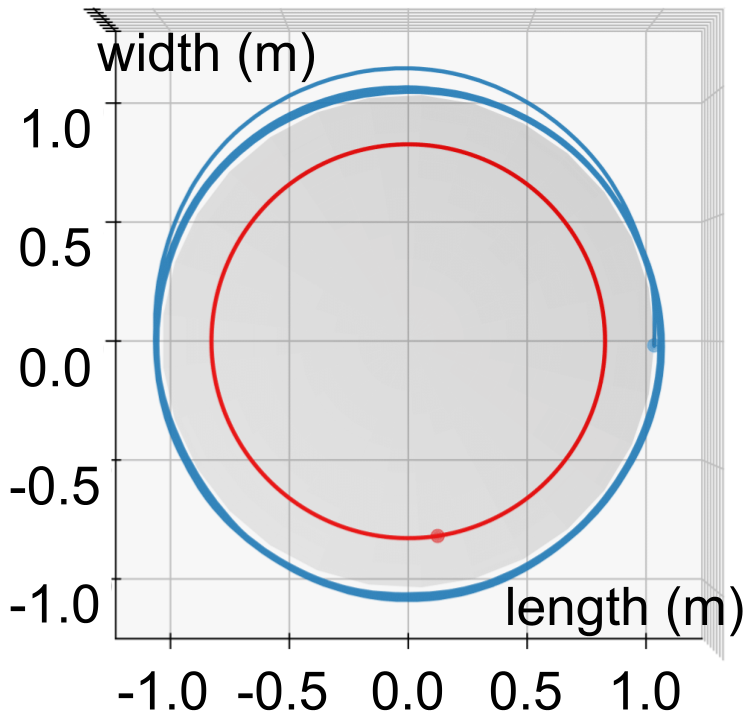}}
    {\includegraphics[width=0.161\textwidth]{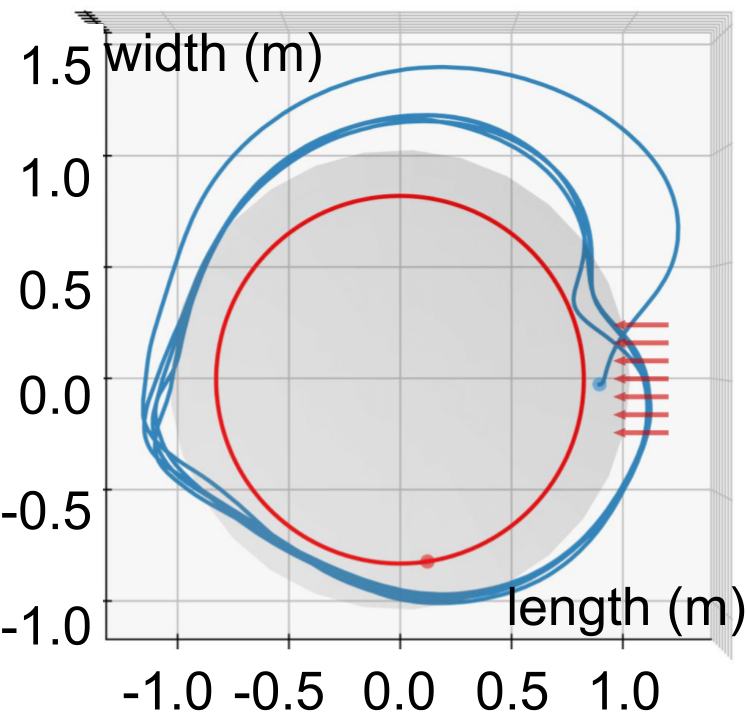}}
    {\includegraphics[width=0.161\textwidth]{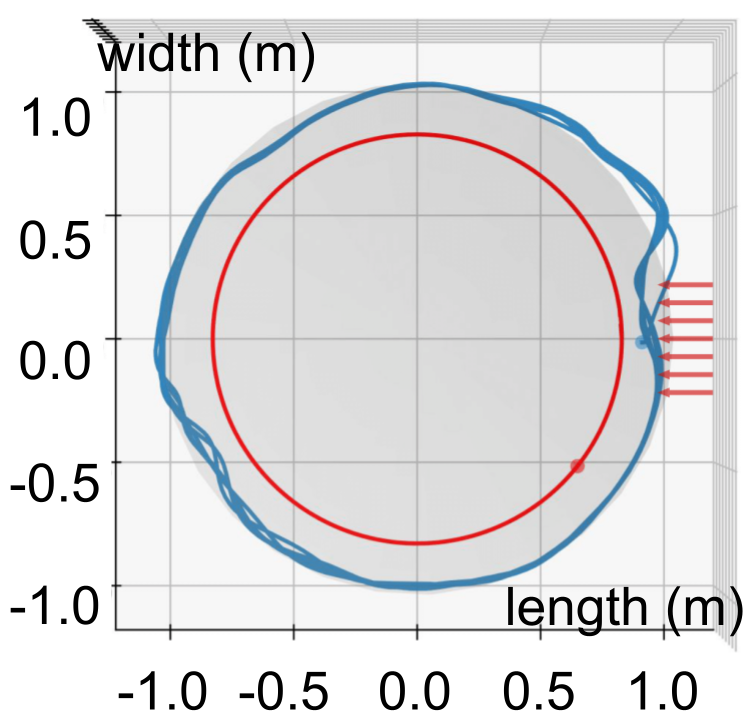}}
    {\includegraphics[width=0.161\textwidth]{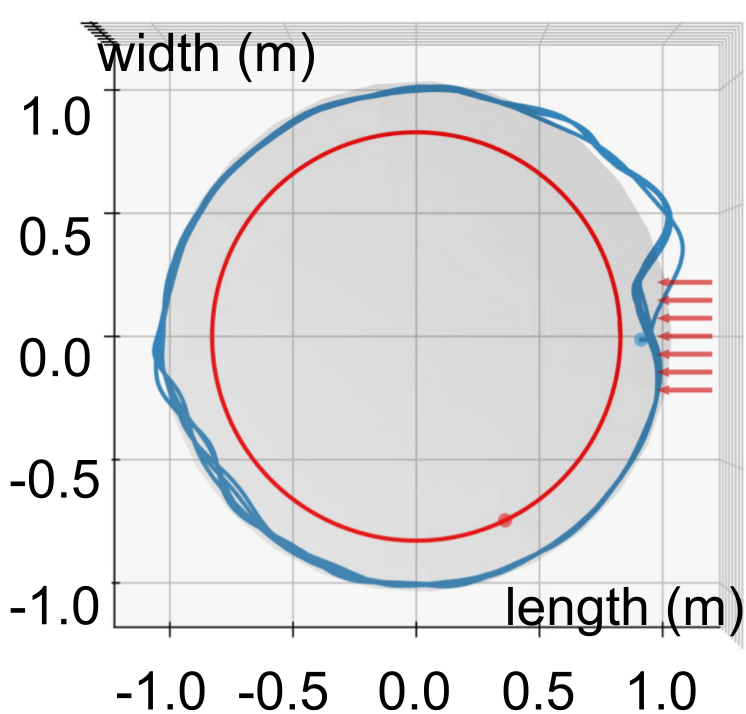}}
    {\includegraphics[width=0.161\textwidth]{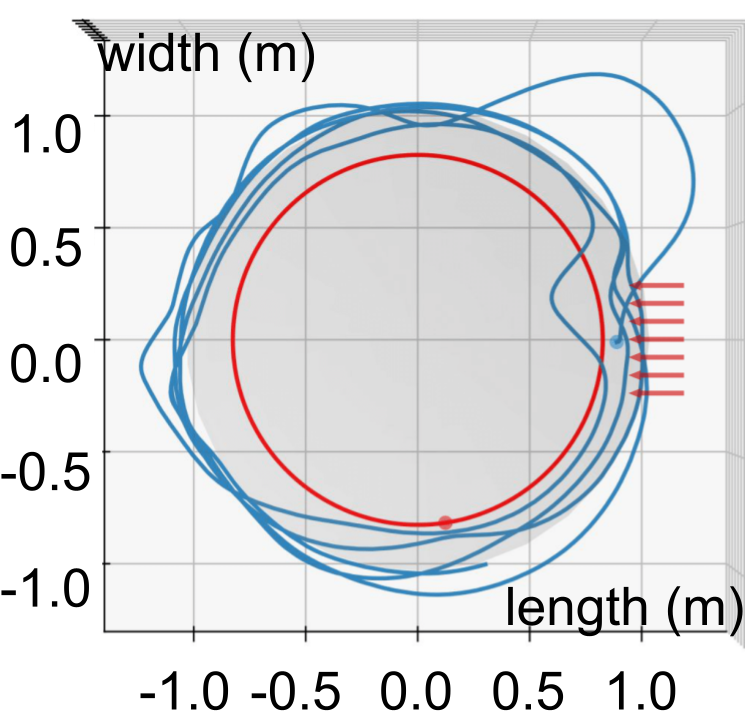}}
    {\includegraphics[width=0.161\textwidth]{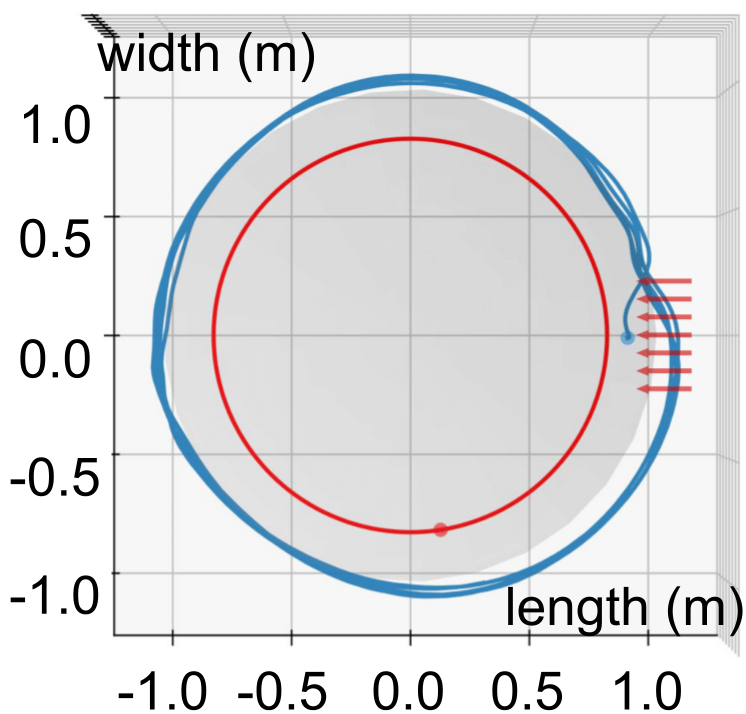}}\\
    \subfloat[HOCBFs - No Wind]{\includegraphics[width=0.161\textwidth]{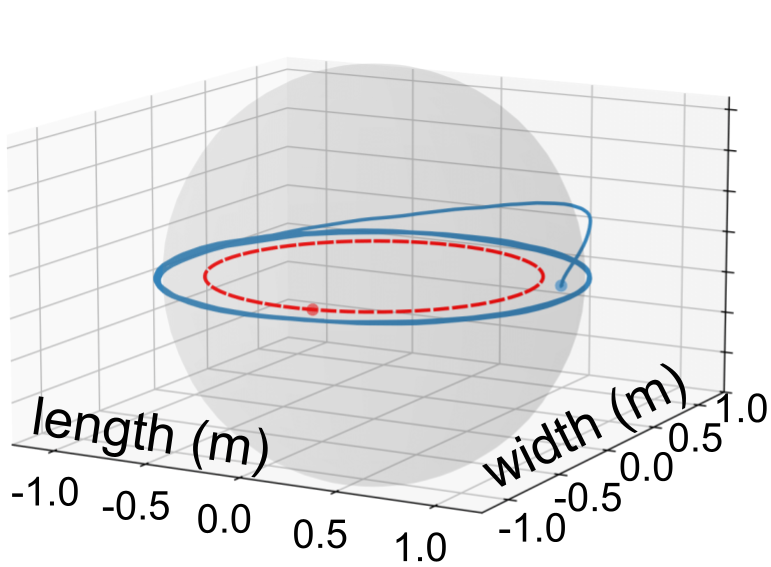}}
    \subfloat[HOCBFs]{\includegraphics[width=0.161\textwidth]{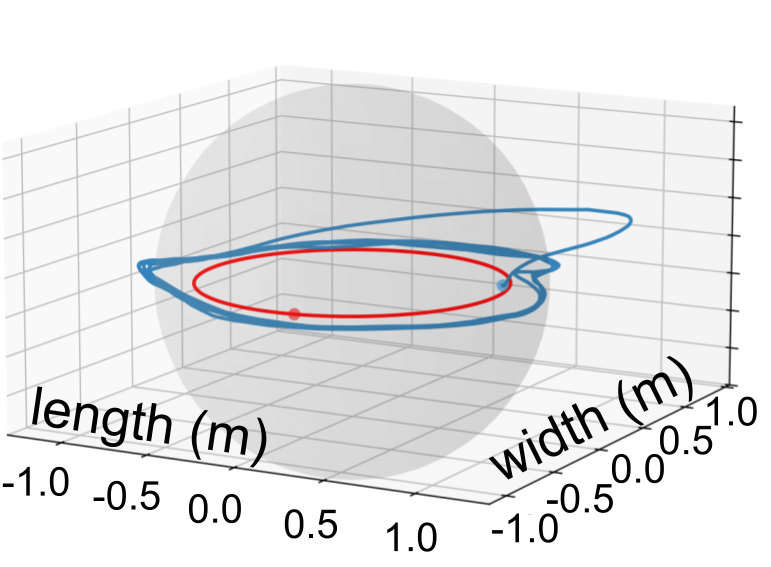}}
    \subfloat[HO-aCBFs State $Y(\mathbf{x})$]{\includegraphics[width=0.161\textwidth]{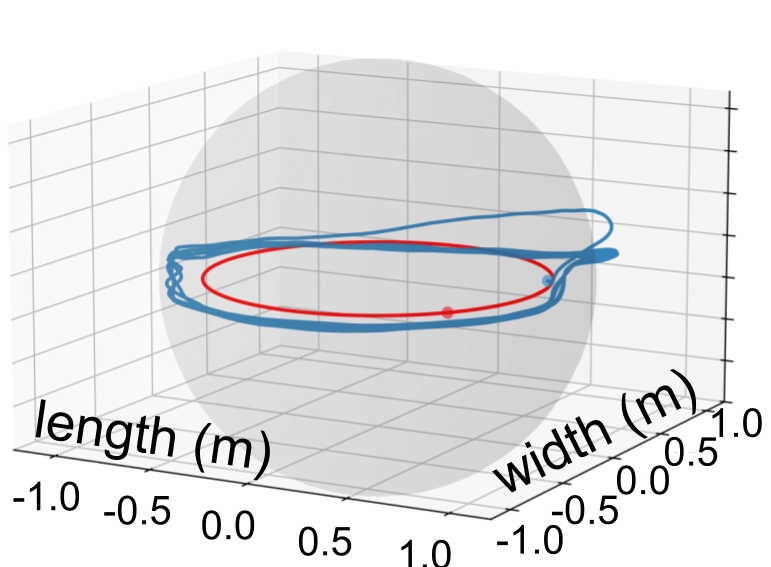}}
    \subfloat[HO-aCBFs Const. $Y(\mathbf{x})$]{\includegraphics[width=0.161\textwidth]{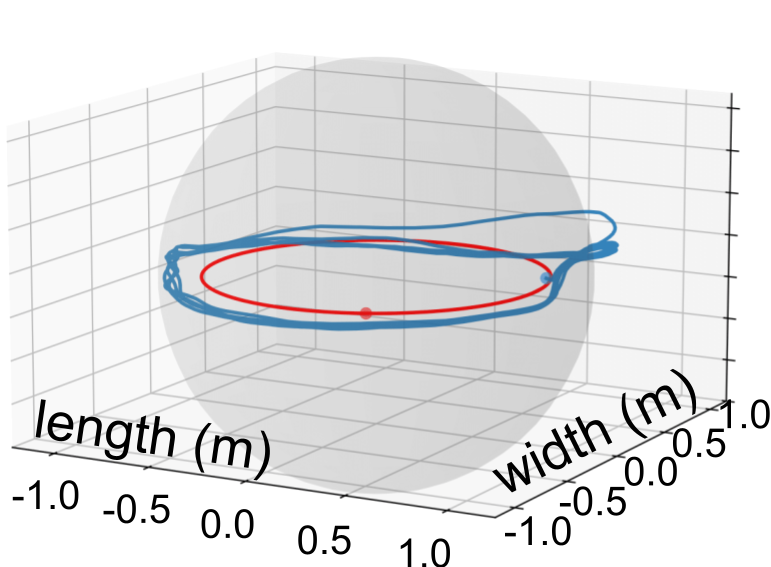}}
    \subfloat[Ours Online]{\includegraphics[width=0.161\textwidth]{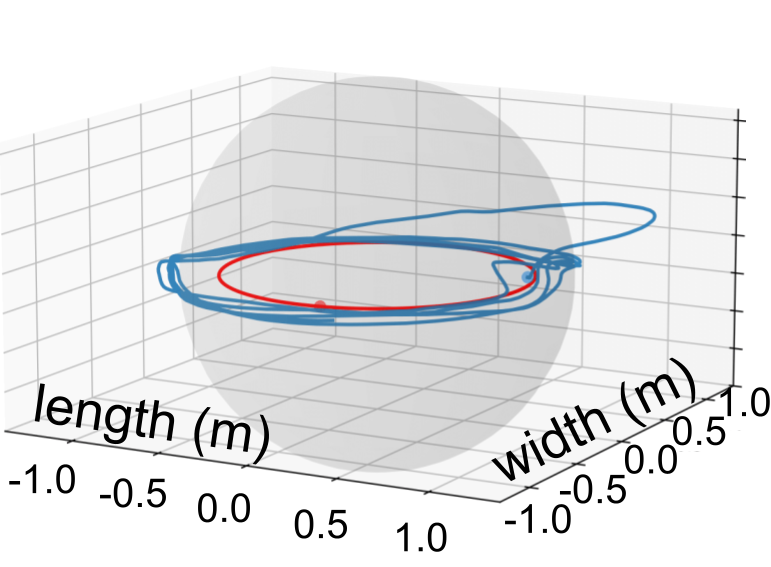}}
    \subfloat[Ours Offline]{\includegraphics[width=0.161\textwidth]{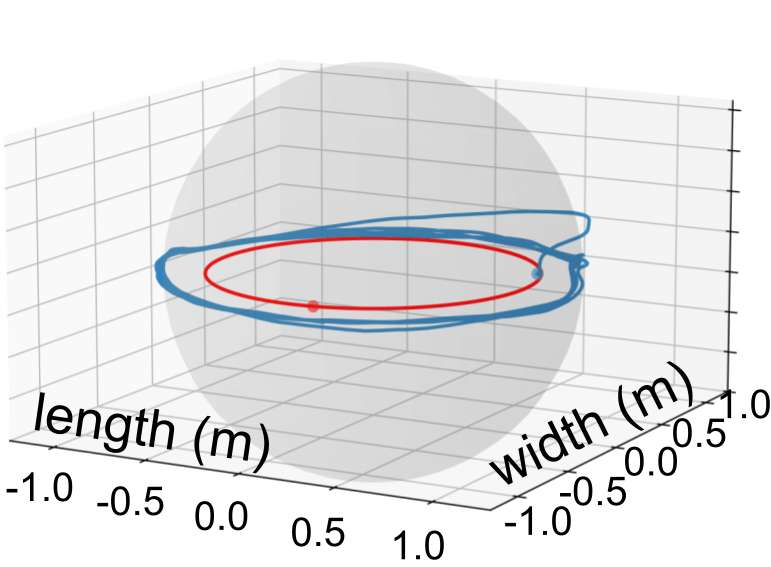}}\\
    {\includegraphics[width=0.99\textwidth]{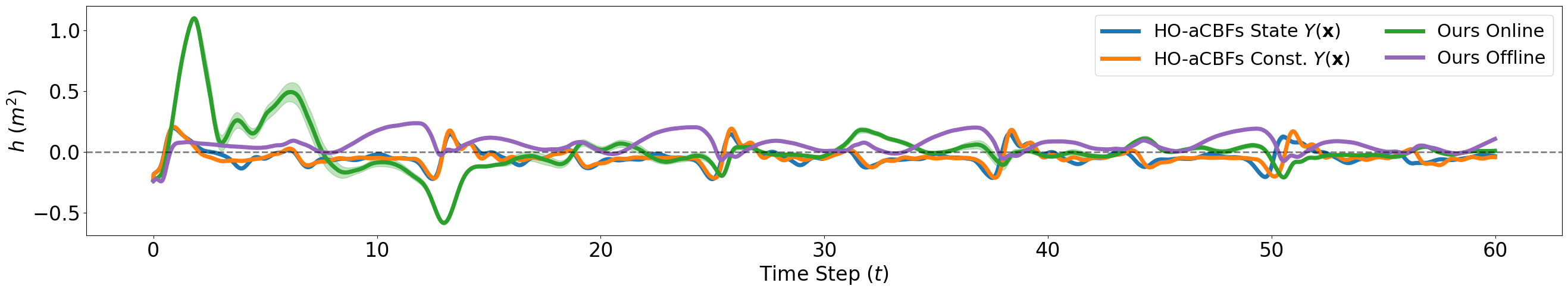}}
    \caption{The top views (first row) and the side views (second row) of $60\unit{s}$ trajectories using different safety controllers in the physical experiments. The red arrows indicate the location and direction of the wind. \newtext{The 60-second time series (third row) of the $h$ value for adaptive controllers.}}
    \label{fig:physical_qualitative}
    \vspace{-1em}
\end{figure*}
\subsection{Simulation Quantitative Results}
We use the following quantitative evaluation criteria:
\begin{itemize}
\item \textit{$h_{\mathrm{min}}$}: the minimum value of $h(\mathbf{x})$  during  execution. 
\item \textit{$h_{\mathrm{neg}}$}: the fractional time for which $h(\mathbf{x})<0$ over execution time. 
\item \textit{Average Distance (Avg.Dist.)}: the average distance to the forbidden region surface over execution time. 
\item \textit{Average settleDistance (Avg.sDist.)}: same as \textit{Avg.Dist.}, excluding the first $20\unit{s}$ in the simulation execution. 
\item \textit{Signed settleDistance Variance (S.sDist.Var.)}: the variance of the signed distance to the forbidden region surface (negative value indicates the robot is inside the forbidden region), excluding the first $20\unit{s}$ in execution.
\end{itemize}
In Table~\ref{table:quantitative}, we record the averaged quantitative results from $10$ simulation trials. \newtext{``CBFs No Res." stands for the experiments using HOCBFs controller without external residuals.} In ``Attractive" and ``Repulsive" experiments, we observe that HO-aCBFs (Tuned), our online and offline controllers can maintain a small \textit{Avg.Dist.} and \textit{Avg.sDist.}, indicating their success in adaptation and safety improvement. However, the HO-aCBF has high $h_{\mathrm{neg}}$, indicating that the HO-aCBF controller failed to estimate a high-fidelity residual to fully recover the safety. Among them, our offline controller has $h_{\mathrm{min}}$ closest to that of HOCBF with no external residuals and $0\%$ of $h_{\mathrm{neg}}$, indicating the highest fidelity of residual estimations. These three controllers all demonstrate small \textit{S.sDist.Var.}, indicating their settled trajectory maintains a stable distance to the forbidden region. All other baselines fail to adapt in ``Attractive" and ``Repulsive" experiments. In ``Time-varying" experiments, only our controllers can adapt. According to \textit{Avg.Dist.} and \textit{Avg.sDist.} metrics, both our online and offline controllers are significantly closer to the safety boundary and have similar results with HOCBF with no external residuals. According to \textit{S.sDist.Var.}, our online and offline controllers settle on trajectories that variate least from the safety boundary, indicating a stable adaptation. Note that even a well-tuned HO-aCBF controller results in large \textit{Avg.Dist.}, \textit{Avg.sDist.} and \textit{S.sDist.Var.}, as depicted in Fig.~\ref{fig:qualitative} (bottom right).  Offline NODE-HO-aCBFs are pre-trained and thus  adapt the residual without delay, resulting in minimal $h_{\mathrm{min}}$. Online NODE-HO-aCBFs have a learning delay, resulting in a slightly larger $h_{\mathrm{min}}$ and \textit{Avg.Dist.} than the offline method, but as it learns online, it adapts better and settles stabler with incoming data and results in a smaller \textit{Avg.sDist.} and \textit{S.sDist.Var.}. \newtext{Note that the metrics of HO-aCBF controller with a large $\Gamma\!=\!1$ differ significantly compared to other methods. As we mentioned in Sec.~\ref{sec:sim_qual_res} that HO-aCBF controller is sensitive to $\Gamma$, and its learnable parameter $\hat{\boldsymbol{\vartheta}}$ fails to converge in the ``Repulsive" and ``Time-varying" experiments, leading to invalid metrics.} 
\begin{figure}[t]
    \centering
    {\includegraphics[width=0.44\textwidth]{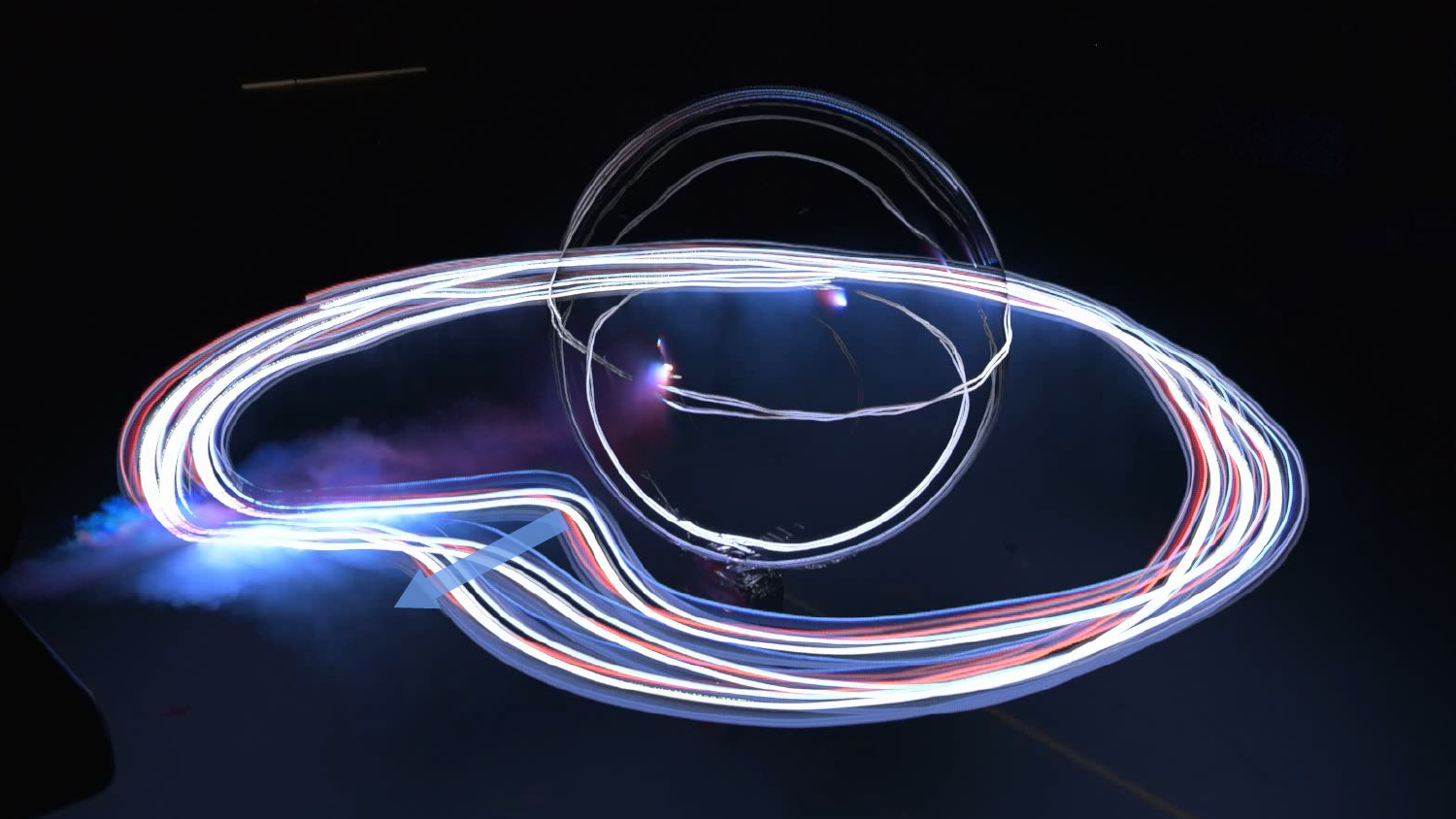}}
    \caption{Long exposure of a nano quadrotor running our online controller under wind turbulence. The blue arrow indicates safety recovery over time.}
    \label{fig:physical_longexp_angle2}
    \vspace{-2em}
\end{figure}
\section{Physical Experiments}
\subsection{Physical Experiment Setup}
The Crazyflie 2.1 is used as the physical platform to verify our controller. The open-source library Crazyswarm~\cite{preiss2017crazyswarm} is used as an interface to send position-velocity-acceleration controls. The nano quadrotor weighs $38\unit{g}$ with a $300\unit{mAh}$ battery. A tower fan is placed horizontally at $0.7\unit{m}$ and $1.16\unit{m}$ in the positive direction of the x-axis from the origin. Its horizontal airflow produces an effective turbulent region of approximately $0.55\unit{m}$ wide, 
which travels $18\unit{km/h}$ (its location and direction are depicted in Fig.~\ref{fig:physical_qualitative}). \newtext{We define a planar circular reference trajectory of radius $0.8\unit{m}$ at height $0.7\unit{m}$ with constant angular velocity $0.5\unit{rad/s}$.} The quadrotor is required to keep a $1\unit{m}$ \newtext{safety }distance from the origin. 
A laptop with Intel i7 is used as the base station, and communication with the Crazyflie is established using Crazyradio PA. We obtain $100\unit{Hz}$ position data from \textsc{Vicon} motion capture, and estimate velocity from filtering. Control inputs are computed and sent to the quadrotor at $100\unit{Hz}$. We increase the number of neurons in each layer to $64$ in both our online and offline controller to estimate the wind turbulence. \newtext{For the online mode, we maintain an FIFO queue of $10\unit{s}$ of online data for training; there is no pre-training.} For the offline mode, we trained on $350\unit{s}$ state-control data collected by executing the HOCBF controller.  A representative physical experiment using our online controller is shown in Fig.~\ref{fig:physical_longexp_angle2}. Note that safety is improved against the wind on the fly. For the HO-aCBF controller, we used the state-informed matrix $Y(\mathbf{x})=[\mathbf{0}_{3\times 3};r_{1},0,0;0,r_{2},0;0,0,r_{3}]$, namely HO-aCBF State $Y(\mathbf{x})$, and a constant matrix $Y(\mathbf{x})=[\mathbf{0}_{3\times 3};\mathbf{I}_{3\times 3}]$, namely HO-aCBF Constant $Y(\mathbf{x})$. 
\subsection{Physical Experiment Results}
We evaluate our controller both qualitatively and quantitatively. In Fig.~\ref{fig:physical_qualitative}, the first $60\unit{s}$ of a representative trajectory for each controller is depicted as the blue curve; the gray area is the forbidden region; the red circle is the reference trajectory; and the red arrows indicate the wind location and direction. Notice that all the baseline methods (Fig.~\ref{fig:physical_qualitative}(b)-(d)) cannot handle the wind turbulence, and converged to an unsafe trajectory. \newtext{As shown by the trajectory and time series in Fig.~\ref{fig:physical_qualitative}, }our online controller improves safety quickly after exposed to wind turbulence, while our pre-trained offline controller significantly improves safety starting from the beginning. 

\begin{figure}[t]
    \centering
    {\includegraphics[width=0.49\textwidth]{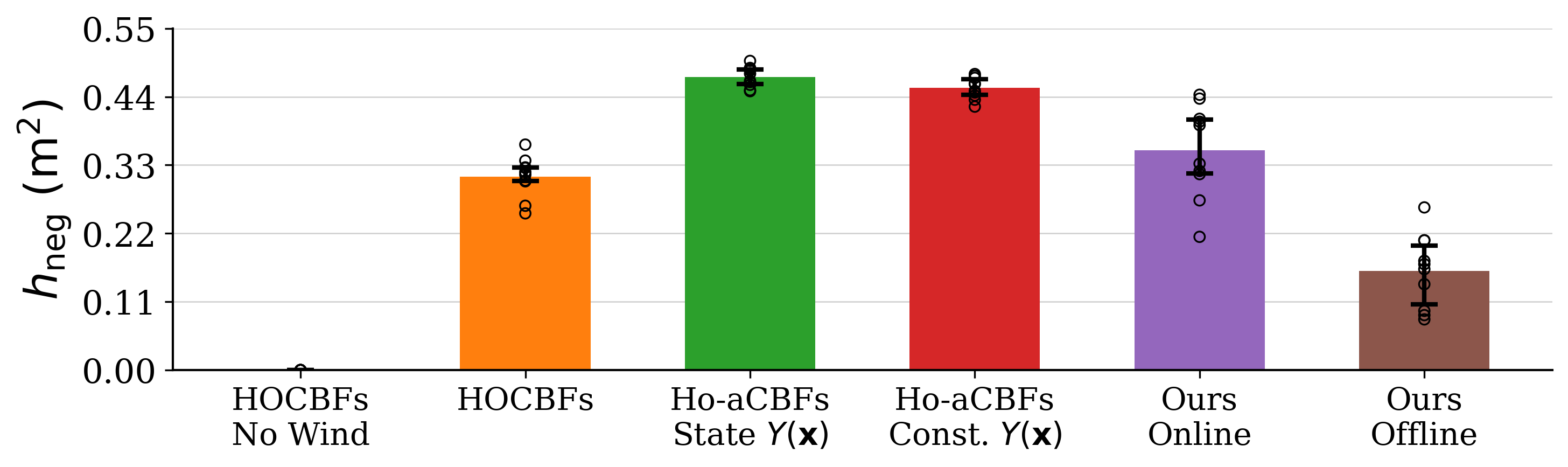}}
    {\includegraphics[width=0.49\textwidth]{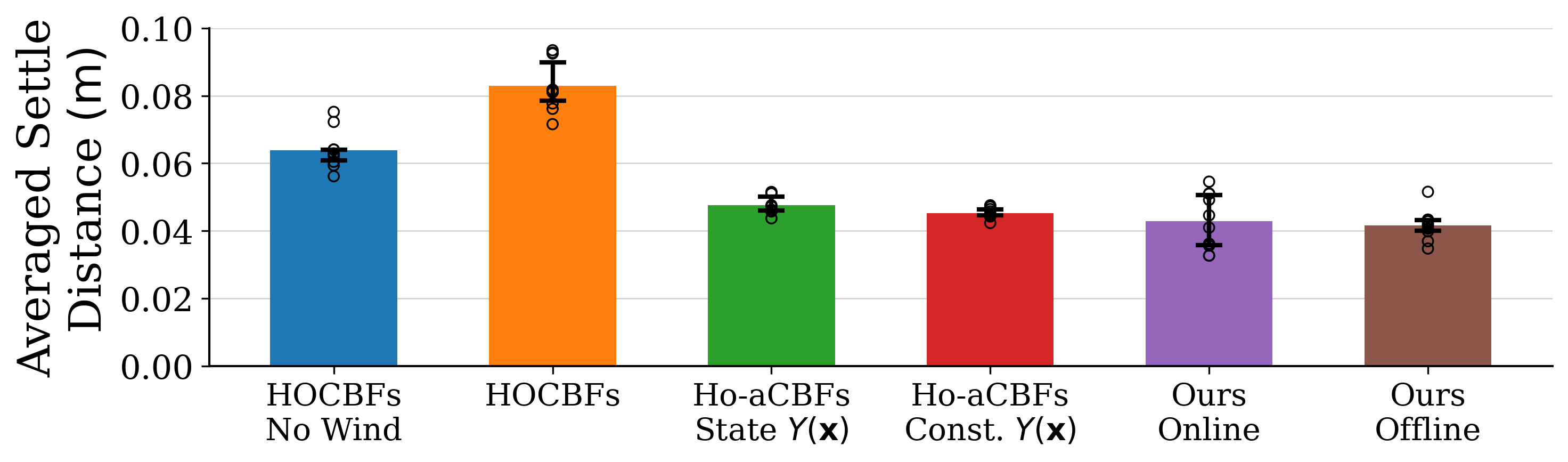}}
    \caption{$h_{\mathrm{neg}}$ and \textit{Avg.sDist.} of baselines and our controllers. The top of the bars denotes the median, while the ends of the error bars represent the $25^{\mathrm{th}}$ and $75^{\mathrm{th}}$ percentiles. 
    The statistics are obtained from $10$ trials.}
    \label{fig:physical_figure}
    \vspace{-2em}
\end{figure}

Table~\ref{table:physical_quantitative} and Fig.~\ref{fig:physical_figure} show the quantitative results in the physical experiments. For the metrics computed on settled trajectory, i.e., \textit{Avg.sDist.} and \textit{S.sDist.Var.}, we exclude the first $60\unit{s}$ of the trajectory. The learning model takes longer to converge due to the sparsity of turbulence data and complex turbulence of the wind. To highlight the adaptation results, we evaluate the trajectory segments where the wind causes significant effects, that is, from $[0,\pi/4]$, and $[\pi, 5\pi/4]$ radian on the x-y plane. 
Our offline approach results in fewer safety violations, as indicated by a smaller $h_{\mathrm{neg}}$, adapts better to turbulence, according to smaller \textit{Avg.Dist.} and \textit{Avg.sDist.}, and settled on a more stable trajectory, according to a smaller \textit{S.sDist.Var.}, compared to HOCBF and HO-aCBF controllers. Note that our offline approach results in smaller \textit{Avg.Dist.} and \textit{Avg.sDist.} even compared to the HOCBF without wind turbulence, as model mismatches exist between a double integrator and a quadrotor model in the HOCBF controller, learning the residuals adapt to such model mismatch. Our online approach has a smaller $h_{\mathrm{neg}}$ than the HO-aCBF controller and is slightly larger than that of HOCBF, as it requires more data to learn the residual. As the online approach sees more data, it adapts turbulence better and settles on a more stable trajectory, as indicated by smaller \textit{Avg.sDist.} and \textit{S.sDist.Var.}, compared to HOCBF and HO-aCBF controllers. 

\begin{table}[h]
\centering
\scalebox{0.9}{
\begin{tabular}{|c|c|c|c|c|c|}
\hline
\multicolumn{1}{|c|}{Method} & \multicolumn{1}{c|}{$h_{\mathrm{min}}$} & \multicolumn{1}{c|}{$h_{\mathrm{neg}}$} & \multicolumn{1}{c|}{Avg.Dist.} & \multicolumn{1}{c|}{Avg.sDist.}  & \multicolumn{1}{c|}{S.sDist.Var.} \\
\hline
\newtext{CBFs - No Wind} & 0.03 & 0.0\% & 6.46e-2 & 6.39e-2  & 1.02e-4  \\
\hline
\newtext{CBFs} & -0.34 & 31.1\% & 9.03e-2 & 8.30e-2  & 7.61e-3 \\
\hline
\newtext{aCBFs State $Y$} & -0.23 & 47.2\% & 4.68e-2 & 4.76e-2  & 3.26e-3 \\
\hline
\newtext{aCBFs Const. $Y$} & \textbf{-0.22} & 45.4\% & 4.49e-2 & 4.53e-2  &  2.90e-3 \\
\hline
\newtext{Ours Online} & -0.55 & 35.3\% & 7.08e-2 & 4.29e-2  & 2.49e-3 \\
\hline
\newtext{Ours Offline} & -0.24 & \textbf{15.9\%} & \textbf{4.44e-2} & \textbf{4.17e-2} & \textbf{1.34e-3} \\
\hline
\end{tabular}
}
\caption{Quantitative comparison between the baseline controllers and our controllers \newtext{(The ``HO-" prefix is dropped in all methods due to space limit)} in the physical experiments. The statistics are averaged across $10$ trials.}
\label{table:physical_quantitative}
\vspace{-2em}
\end{table}
\section{Conclusion}
In this work, we present a learning-enhanced high-order adaptive control barrier function that improves safety online under time-varying complex model perturbations. We use a hybrid structure that combines a nominal model with a neural network residual model, using neural ODEs. Due to its hybrid design, our controller is data-efficient and expressive, making it suitable for adapting complex, time-varying residuals online. We demonstrate its efficacy across different simulated residuals and a representative physical experiment, benchmarked with baseline HOCBFs~\cite{xiao2021high} and HO-aCBFs~\cite{isaly2021adaptive, cohen2022high} (with different hyperparameters). A $38\unit{g}$ nano quadrotor, deployed with our online NODE-HO-aCBF controller, can quickly improve its safety performance in $18\unit{km/h}$ turbulent wind. Current large-scale swarm trajectory planning algorithms, such as~\cite{pan2024hierarchical, pan2025hierarchical, luis2020online}, lack an adaptive controller that improves safety and stability in close-proximity flight. For future work, we aim to provide probabilistic safety guarantees for the controller proposed in this work and to improve the safety and stability of close-proximity trajectory planning for large-scale swarms. 



\bibliographystyle{IEEEtran}
\bibliography{IEEEabrv, refs}

\end{document}